\documentclass{article}

\usepackage{microtype}
\usepackage{graphicx}
\usepackage{subfigure}
\usepackage{booktabs} 
\usepackage{caption}

\usepackage{thmtools} 
\usepackage{thm-restate}

\usepackage{hyperref}
\usepackage{color}
\usepackage[table]{xcolor}

\usepackage[accepted]{icml2025}

\usepackage{amsmath}
\usepackage{amssymb}
\usepackage{mathtools}
\usepackage{amsthm}
\usepackage{bm}

\definecolor{Gray}{gray}{0.85}

\newcolumntype{a}{>{\columncolor{Gray}}c}

\usepackage[capitalize,noabbrev]{cleveref}

\theoremstyle{plain}

\theoremstyle{definition}
\newtheorem{definition}{Definition}

\theoremstyle{remark}

\newcommand{\method}{\textsc{CATS}}
\newcommand{\mtdfunc}{\phi}
\newcommand{\ie}{i.e.}
\newcommand{\mapping}[2]{#1 $\rightarrow$ #2}
\newcommand{\mtx}[1]{\mathbf{#1}}

\newcommand{\adpf}{f_{\texttt{\method}}}
\newcommand{\fctHead}{g_\texttt{{f}}}
\newcommand{\cor}{\mathrm{corr}}

\icmltitlerunning{CATS: Mitigating Correlation Shift for Multivariate Time Series Classification}

\begin{document}

\twocolumn[
\icmltitle{CATS: Mitigating Correlation Shift \\ for Multivariate Time Series Classification}




\icmlsetsymbol{equal}{*}

\begin{icmlauthorlist}
\icmlauthor{Xiao Lin}{illinois}
\icmlauthor{Zhichen Zeng}{illinois}
\icmlauthor{Tianxin Wei}{illinois}
\icmlauthor{Zhining Liu}{illinois}
\icmlauthor{Yuzhong chen}{visa}
\icmlauthor{Hanghang Tong}{illinois}
\end{icmlauthorlist}

\icmlaffiliation{illinois}{Siebel School of Computing and Data Science, University of Illinois Urbana-Champaign, Illinois, US.}
\icmlaffiliation{visa}{Visa Research, Foster City, CA}

\icmlcorrespondingauthor{Xiao Lin}{xiaol13@illinois.edu}
\icmlcorrespondingauthor{Hanghang Tong}{htong@illinois.edu}

\icmlkeywords{Machine Learning, ICML}

\vskip 0.3in
]



\printAffiliationsAndNotice{\icmlEqualContribution}
\begin{abstract}
    Unsupervised Domain Adaptation (UDA) leverages labeled source data to train models for unlabeled target data. Given the prevalence of multivariate time series (MTS) data across various domains, the UDA task for MTS classification has emerged as a critical challenge. 
    However, for MTS data, correlations between variables often vary across domains, whereas most existing UDA works for MTS classification have overlooked this essential characteristic. 
    To bridge this gap, we introduce a novel domain shift, {\em correlation shift}, measuring domain differences in multivariate correlation. To mitigate correlation shift, we propose a scalable and parameter-efficient \underline{C}orrelation \underline{A}dapter for M\underline{TS} (\method{}). Designed as a plug-and-play technique compatible with various Transformer variants, \method{} employs temporal convolution to capture local temporal patterns and a graph attention module to model the changing multivariate correlation. 
    The adapter reweights the target correlations to align the source correlations with a theoretically guaranteed precision.
    A correlation alignment loss is further proposed to mitigate correlation shift, bypassing the alignment challenge from the non-i.i.d. nature of MTS data.
    Extensive experiments on four real-world datasets demonstrate that (1) compared with vanilla Transformer-based models, \method{} increases over $10\%$ average accuracy while only adding around $1\%$ parameters, and (2) all Transformer variants equipped with \method{} either reach or surpass state-of-the-art baselines.

\end{abstract}
\addtocontents{toc}{\protect\setcounter{tocdepth}{-1}}
\section{Introduction}

Multivariate time series classification (MTS) is a fundamental task with applications spanning diverse fields, including finance \cite{zhao2023stock, shahi2020stock, mondal2014study, lebaron1999time}, healthcare~\cite{zeger2006time, touloumi2004analysis, dockery1996epidemiology, bernal2017interrupted}, climate science~ \cite{yoo2020time, ghil2002advanced, belda2014climate, baranowski2015multifractal}, transportation  \cite{rezaei2019deep, montazerishatoori2020detection, vu2018time} and power systems \cite{hoffmann2020review, futterer2017application, susto2018time}. Recently, deep learning models \cite{vaswani2017attention, liu2023itransformer, wu2022timesnet} have demonstrated remarkable capability in capturing temporal dependencies within time series data, showcasing significant promise in numerous applications.

However, the deployment of these models often encounters a critical challenge: {\em domain shifts} \cite{koh2021wilds, luo2019taking, zhang2013domain} between the labeled source domain data during training and the target domain data during testing. The domain shift often leads to a notable degradation in model performance on the target domain. Moreover, obtaining labels for the test data is quite hard in most real-world scenarios \cite{ganin2016domain, long2015learning}.
As a result, the UDA problem on MTS \cite{he2023domain, wilson2020multi} has emerged as a critical research area \cite{he2023domain, wilson2020multi}, aiming to leverage the labeled source domain data to enhance the model performance on the unlabeled target domain.

Previous studies on UDA for MTS primarily focus on learning domain-invariant features through adversarial training \cite{wilson2020multi, wilson2023calda}, contrastive learning \cite{ozyurt2022contrastive} or divergence metrics \cite{he2023domain, cai2021time}. However, these approaches have notable limitations. 
(1)  Model architecture perspective: \textbf{most prior works cannot be easily adapted to Transformer-based architectures}, which have demonstrated state-of-the-art performance in MTS analysis \cite{wu2022timesnet, liu2023itransformer}. Prior works are fundamentally designed for convolutional neural networks (CNNs) \cite{liu2021adversarial, wilson2020multi, he2023domain} or recurrent neural networks (RNNs) \cite{li2022time, wilson2020survey}, with their training strategies and loss functions tightly coupled to these architectures, resulting in the lack of flexibility to Transformer models.
(2) Data distribution perspective: \textbf{prior work overlooks a key characteristic of MTS: varying multivariate correlation}. 
We observe consistent and significant shifts in inter-variable dependencies, i.e., correlations, across domains, and we term this new type of domain shift as \textbf{correlation shift}.
However, most of the existing UDA for MTS approaches overlook this challenge, leaving significant room for improvement by addressing the correlation shifts.

To overcome these limitations, we seek to align the correlation distributions between domains from both the model architecture and the training objective perspectives. At the model level, we propose a scalable and parameter-efficient adapter, termed \underline{C}orrelation \underline{A}dapter for Multivariate \underline{T}ime \underline{S}eries (\method{}). Specifically, to capture temporal dependencies, \method{} employs depthwise convolutions along the temporal dimension for both down-projection and up-projection. Compared to traditional adapters that rely on linear matrices, convolutions demonstrate superior capability in capturing local temporal patterns. 
Building on this, CATS incorporates a Graph Attention Network (GAT) to adaptively reweight inter-variable dependencies in the hidden layers. Theoretically, proper reweighting can align the correlation distributions across domains, thus mitigating correlation shift. To adapt CATS to the unlabeled target domain, we introduce a novel correlation alignment loss. This loss function not only effectively reduces correlation shift but also circumvents the limitations of divergence metrics, which often fail on the non-i.i.d. (non-independent and identically distributed) nature of time series data. By integrating CATS with this tailored loss function, we present a more effective and efficient solution for unsupervised domain adaptation in multivariate time series. 

In summary, our main contributions are as follows:
\begin{itemize}
    \item \textbf{Problem.} We are the first to identify and empirically validate the phenomenon of correlation shift in MTS data. As all prior works have overlooked the discrepancy in multivariate correlation across domains, correlation shift introduces a new perspective for addressing UDA challenges in MTS.
    \item \textbf{Model.} We propose the first scalable and parameter-efficient MTS adapter, \method{}, designed specifically for large-scale time series Transformers. Empirically and theoretically, \method{} effectively mitigates correlation shifts while capturing local temporal patterns for classification.
    \item \textbf{Training objective.} We introduce a novel correlation alignment loss, which directly addresses correlation shift and circumvents the alignment challenge posed by the non-i.i.d. MTS data.
    \item \textbf{Evaluation.} We conduct extensive experiments on four real-world time series domain adaptation datasets. The results demonstrate that \method{} consistently enhances the performance of Transformer variants, achieving a $10\%+$ average accuracy improvement, even under large domain shifts. Furthermore, Transformer variants equipped with \method{} outperform SOTA baselines by around 4\% average accuracy, showcasing the superb effectiveness and adaptability of \method{}.
\end{itemize}

\vspace{-1mm}
\section{Preliminaries} \label{sec:prelim}

\noindent\textbf{Multivariate time series classification.}
In the task of multivariate time series (MTS) classification, the dataset is comprised of a collection of time series samples along with their corresponding labels, denoted as $\mathcal{D} = \{ (\mathbf{X}_i, y_i)\}_{i=1}^n$ with $n$ being the sample number. Here, the $i$-th sample $\mathbf{X}_i \in \mathbb{R}^{D\times T}$ represents an individual time series that contains readouts of $D$ observations over $T$ time points, and $y_i$ is the associated label. In this paper, we use $\mathbf{X}[j]$ to represent the $j$-th variable of the sample $\mathbf{X}$.

\noindent\textbf{Adapters for large models.} 
Recently, large-scale Transformers have achieved remarkable success across various fields, including natural language processing \cite{vaswani2017attention, devlin2018bert, brown2020language}, computer vision \cite{radford2021learning, alexey2020image}, and time series analysis \cite{liu2023itransformer, wu2022timesnet}. However, due to the massive number of parameters, it is highly impractical to fine-tune a pretrained Transformer for every downstream task. To address this challenge, numerous parameter-efficient fine-tuning (PEFT) methods \cite{han2024parameter, hu2021lora, xu2023parameter} have been proposed. Among these, {\em adapters} have garnered significant attention because of their capabilities of transferring rich internal knowledge of the pretrained model to downstream tasks at the cost of a small number of additional parameters \cite{hu2023llm}.

Given the high similarity between the objectives of adapters and UDA, many studies \cite{zhang2021unsupervised, malik2023udapter} have leveraged adapters to transfer knowledge learned from the source domain to the target domain. Such domain adapters are embedded between two consecutive Transformer blocks to adapt the model's learned representations to the target domain. Mathematically, these adapters can be expressed as: 
\begin{equation} \label{eq:typ_adp}
    \begin{aligned}
        \mtx{H}^{(k)}_O = \mtx{H}_I^{(k)} + \sigma(\mtx{H}_I^{(k)} \mtx{W}_{\downarrow}^{(k)}) \mtx{W}_{\uparrow}^{(k)} 
    \end{aligned}
\end{equation}
where $\sigma(\cdot)$ represents the activation function, and $\mtx{W}_{\downarrow}^{(k)} \in \mathbb{R}^{T \times r}$ and $\mtx{W}_{\uparrow}^{(k)} \in \mathbb{R}^{r \times T}$ are the two linear matrices for down-projection and up-projection with $r$ being a small hyperparameter. Here,  $\mtx{H}^{(k)}_I$ is the input of the $k$-th adapter block, and $\mtx{H}^{(k)}_O$ is the output of the $k$-th adapter, \ie, the input of the $(k+1)$-th Transformer block. 

\noindent\textbf{Unsupervised domain adaptation.}
The goal of UDA is to leverage information from a labeled source domain $\mathcal{D}_s=\{(\mathbf{X}_{i,s}, y_{i,s})\}_{i=1}^{n_s}$ to enhance the model’s understanding of an unlabeled target domain $\mathcal{D}_t=\{\mathbf{X}_{i, t}\}_{i=1}^{n_t}$.  Generally, source and target samples are independently sampled from their respective distributions, \ie, $\mathcal{D}_s \sim \mathcal{P}_s(\mathbf{X}_s, y_s)$ and $\mathcal{D}_t \sim \mathcal{P}_t(\mathbf{X}_t, y_t)$. However, these distributions often exhibit significant shifts. There are two widely studied shifts: feature shift \cite{zhang2013domain} and label shift \cite{azizzadenesheli2019regularized}.
Specifically, feature shift occurs when the distribution of features changes across domains, while the relationship between features and labels remains consistent. In contrast, label shift arises when the label distributions differ between domains, even if the feature distributions are similar.

\vspace{-1mm}
\section{Correlation shift}

Although label shift and feature shift are the two most commonly analyzed types of domain shifts in UDA tasks, focusing solely on these shifts is insufficient for MTS classification. A key characteristic of MTS is the interaction between different variables, such as the interplay between blood glucose levels and insulin in the human body \cite{basu2009effects, wang2018delay}. Correlation effectively models this inter-variable dependencies, thus making it central to many statistical and deep learning models for MTS \cite{box2015time, bollerslev1990modelling, wu2021autoformer}.

However, prior UDA methods for MTS, either implicitly or explicitly, have largely overlooked this crucial property. To address this gap, we introduce a novel shift tailored specifically for MTS: \textbf{correlation shift}.
\begin{definition}[Correlation shift]
    Suppose the source multivariate data $\mtx{X}_s \in \mathbb{R}^{D\times T}$ and the target multivaraite data $\mtx{X}_t\in \mathbb{R}^{D\times T}$ follow the source distribution $\mathcal{P}_s$ and the target distribution $\mathcal{P}_t$, respectively. Here, $D$ denotes the number of variables and $T$ represents the feature dimension. Then, \textbf{correlation shift} occurs when the multivariate correlations between the source and target domains differ, formally defined as:
    \begin{align}
        &\mathrm{Corr}(\mathbf X_s) \neq \mathrm{Corr}(\mathbf X_t)
    \end{align}
    where the correlation structure $\mathrm{Corr}\left(\cdot\right)$ is given by:
    \begin{equation}\label{eq:corr_dfn}
        \begin{aligned}
            &\mathrm{Corr}(\mathbf X) := \mathrm{diag}(\bm{\Sigma})^{-1/2}\bm{\Sigma}\mathrm{diag}(\bm{\Sigma})^{-1/2} \\
        &\mathbf{\Sigma}  = \mathbb E_{\mathbf X\sim \mathcal{P}}\left[ (\mathbf X - \mathbb E \mathbf X)(\mathbf X-\mathbb E \mathbf X)^T \right] 
        \end{aligned}
    \end{equation}
\end{definition}

\vspace{-1mm}
This phenomenon naturally arises from discrepancies in inter-variable dependencies across domains.
A practical example of the correlation shift can be observed in healthcare analytics. 
For example, in non-diabetic individuals, there is typically a synchronous peak in blood glucose and insulin levels following sugar intake while in diabetic patients, the increase in insulin occurs with a noticeable delay after the peak in blood glucose \cite{basu2009effects, wang2018delay}. This delay represents a clear correlation shift when considering blood glucose and sugar intake as two interacting variables.
Another widely-existing example comes from the weather data. Extensive studies \cite{draper2004simultaneous, weissman2002effects, back2005relationship} have shown that the relationship between wind speed and precipitation varies geographically and this relationship tends to be significantly stronger in humid regions compared to arid areas. The widespread occurrence of correlation shifts impacts the transferability of learned representations, ultimately leading to  deteriorated performance on target domains

\begin{minipage}{0.49\linewidth}
  \includegraphics[width=\linewidth, trim=10 15 0 0, clip]{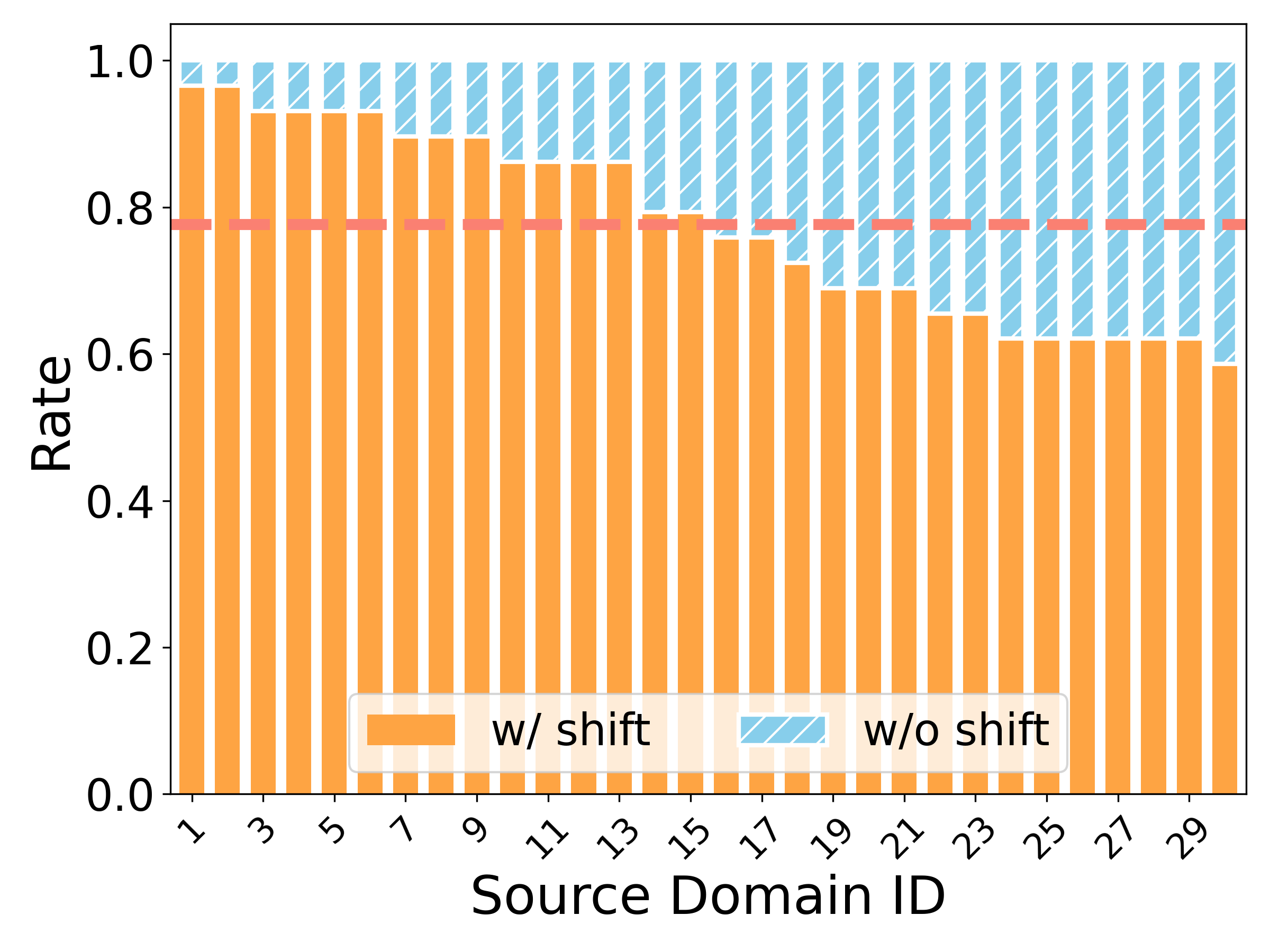}
\end{minipage}
\hfill
\begin{minipage}{0.49\linewidth}
  \captionof{figure}{Rates of target domains with correlation shifts per source domain. The x-axis represents the source domain index while the y-axis indicates the rate of correlation shifts among the rest 29 domains. The red line marks the average rate of 78\%.} 
  \label{fig:p_value_x_shift}
\end{minipage}


To further validate the universality of correlation shifts, we conduct an empirical analysis on a real-world Human Activity Recognition (HAR) dataset \cite{anguita2013public}, to demonstrate the discrepancy in the correlation across different domains.
Specifically, we iterate through the 30 domains in HAR, treating each domain as the source domain while considering the remaining 29 domains as target domains.
For each source-target domain pair, we apply the Mann-Whitney U test \cite{mcknight2010mann}, a non-parametric hypothesis testing method, to determine whether there is a significant correlation shift between the source and the target domains.
A detailed explanation for this correlation shift test is provided in Appendix \ref{appdix:emp_cor_shift}. 
We compute the rate of target domains suffering from significant correlation shifts for each source domain, and the results are shown in Figure \ref{fig:p_value_x_shift}, where the orange bars indicate the rate of target domains with significant correlation shifts. 
The red dashed line in Figure \ref{fig:p_value_x_shift} marks the average rate of correlation shift, which is 78\%. These findings provide clear evidence that correlation shifts are indeed prevalent in MTS datasets, thereby calling for solutions to mitigate them.

\vspace{-1mm}
\section{Methodology}

\begin{figure*}[htbp]
    \centering
    \resizebox{\linewidth}{!}{
        \includegraphics[width=0.5\linewidth]{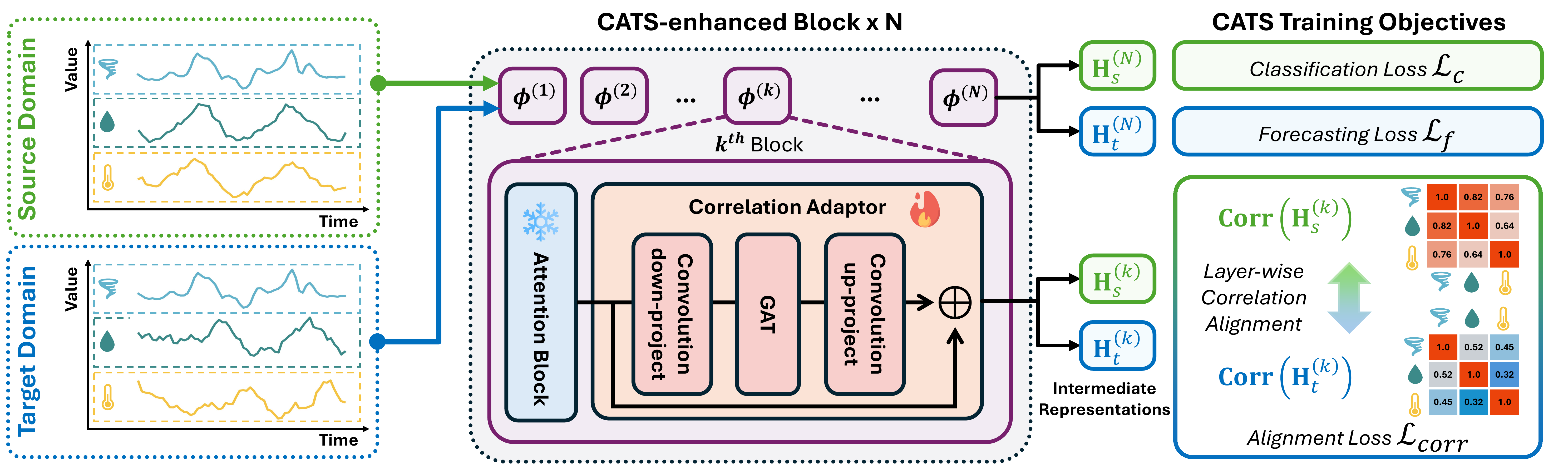}
    }
    \caption{The main framework of \method{}. \method{} is integrated after each attention block of any Transformer variant, with only \method{} trained and the backbone frozen. The training objective involves three loss functions: (1) classification loss on the labeled source domain, (2) forecasting loss on the unlabeled target domain, and (3) layer-wise correlation alignment loss to align these two domains.}
    \label{fig:pipeline}
\end{figure*}

In this section, we introduce our solution to mitigate correlation shift. We first propose \method{} in Section \ref{sec:adapter}, which demonstrates superior representation learning capabilities for MTS compared to traditional adapters.
By reweighting multivariate correlation, \method{} enjoys theoretical guarantees for mitigating correlation shift. In Section \ref{sec:train}, we propose a novel training objective for \method{} on unlabeled target domains, centered on the correlation alignment loss to address correlation shift effectively. The overall framework of \method{} is visualized in Figure \ref{fig:pipeline}.

\vspace{-1mm}
\subsection{\method: Temporal-aware Correlation Adapter} \label{sec:adapter}
MTS data exhibit two prominent properties: temporal dependencies and inter-variable dependencies. To model these properties, we first introduce up-project and down-project modules for time series adapters in Section \ref{sec:TDC} to capture temporal dependencies effectively. In Section \ref{sec:gat}, we further propose a reweighting module to adaptively refine inter-variable dependencies, thereby mitigating the impact of correlation shift. By integrating these components, we propose \method{} which effectively enhances the model's adaptability in domain adaptation tasks involving multivariate time series classification. 

\vspace{-1mm}
\subsubsection{Temporal Project via Convolution} \label{sec:TDC}

As discussed in Section \ref{sec:prelim}, adapters hold significant potential for addressing domain adaptation challenges in Transformer models.
However, these previous adapters often rely on the assumption that the data are i.i.d. (independent and identically distributed), which does not hold for MTS. A key property of MTS is that temporally adjacent data points often exhibit strong similarity. However, the use of linear matrices in existing adapters fails to capture this local similarity, leading to noticeable declines in performance. Inspired by temporal convolution network (TCN) \cite{fan2023parallel, farha2019ms, hewage2020temporal}, we posit that convolutions on temporal dimension could better leverage local similarity on MTS, thus serving as a better substitute as project layer, compared with linear matrices in adapters from Eq. \eqref{eq:typ_adp}.

One potential drawback of using convolutions along the temporal dimension is the increase of the number of trainable parameters. On previous adapters in Eq.~\eqref{eq:typ_adp}, the parameter complexity of the linear matrices are $\mathcal{O}(T \times r)$. In contrast, convolutions have a parameter complexity of $\mathcal{O}(D^2 \times r)$, where $r$ is the kernel size. When the hidden layer dimension 
$D$ approaches or exceeds the time length $T$, the number of trainable parameters for convolutions can become quite large. To address this issue, we adopt depthwise convolutions \cite{chollet2017xception}, where each variable is convolved with its own kernel. This approach reduces the parameter complexity to $\mathcal{O}(D \times r)$, significantly improving efficiency. Note that depthwise convolutions ignore the multivariate correlation. We will address this issue in Section \ref{sec:gat}.

\begin{minipage}{0.47\linewidth}
  \includegraphics[width=\linewidth]{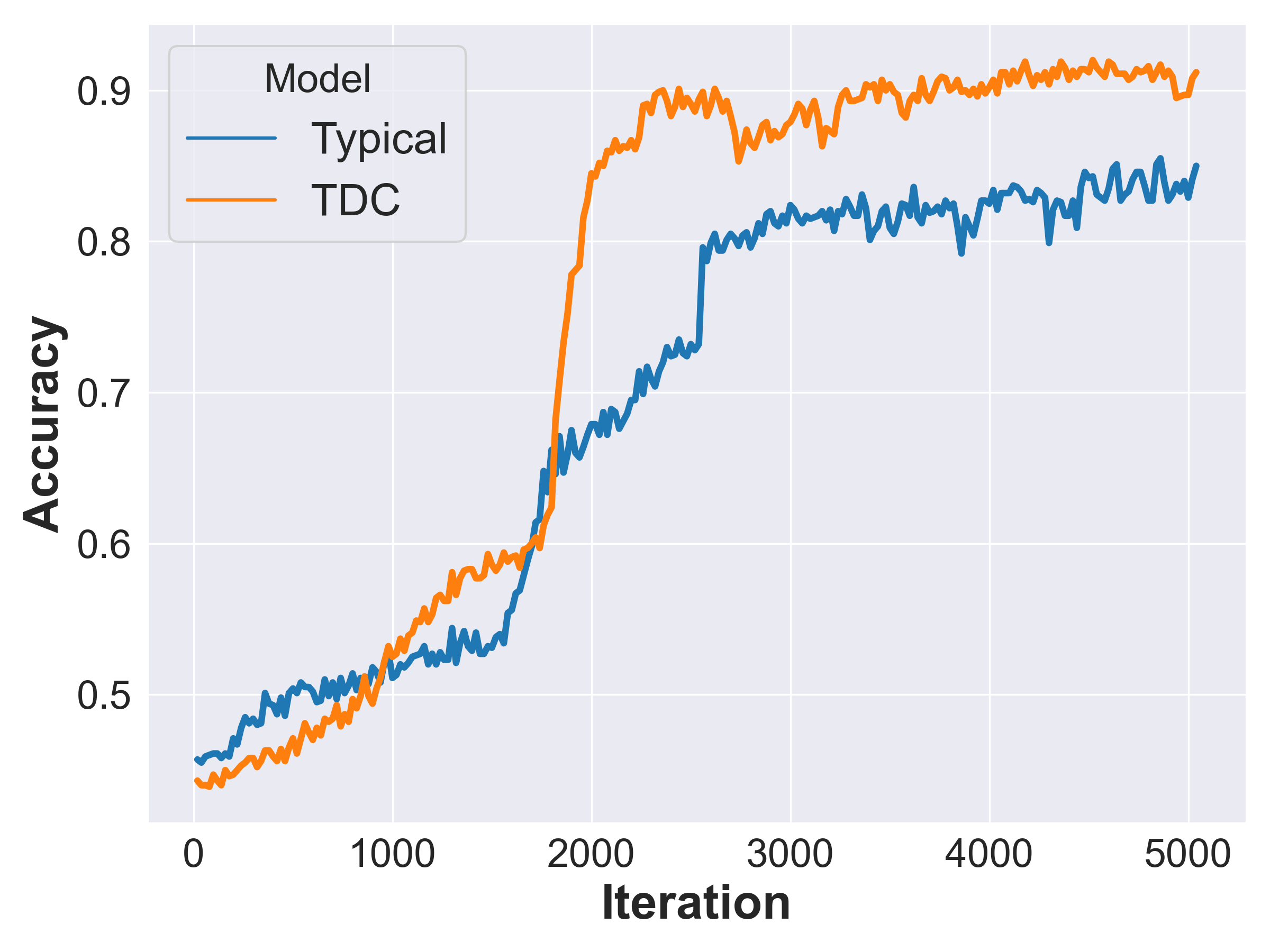}
\end{minipage}
\hfill
\begin{minipage}{0.5\linewidth}
  \captionof{figure}{The accuracy comparison on HAR dataset between the typical adapter and the TDC-based adapter. With the backbone (TimesNet) pretrained on the domain 1, both adapters are trained on the domain 10.} 
  \label{fig:adp_compare}
\end{minipage}

To empirically validate the superiority of temporal depthwise convolutions (TDC), we compare the performance between a typical adapter in Eq. \eqref{eq:typ_adp} and a TDC-based adapter which only uses temporal depthwise convolutions instead of the linear layers. 
Specifically, given a backbone pretrained on the source dataset, we will train these two adapters on the target domain. Note that the task is designed to assess the representation learning ability of different adapters on time series rather than focusing solely on domain adaptation. Therefore, we use the labels from the target domain when training adapters. A detailed experimental setting is provided in Appendix \ref{appdix:compare_tdc_linear}. The experiment results, illustrated in Figure \ref{fig:adp_compare}, indicate that the TDC-based adapter demonstrates significantly higher accuracy (around 10\% improvement) after both adapters converge. This clearly supports the premise that, for time series data, temporal convolutions are a superior alternative to traditional linear matrices.


\vspace{-2mm}
\subsubsection{Correlation Alignment via Reweighting} \label{sec:gat}

In this section, our objective is to identify an effective reweighting module to mitigate the correlation shift. Since correlation shift arises from discrepancies in multivariate correlations between the source and target domains, a natural approach to addressing it is to adaptively reweight the correlations in the target domain. Interestingly, in the case of Gaussian variables, we prove that a simple linear mapping \footnote{Although adapters in Eq. \eqref{eq:typ_adp} contains linear matrices, it is not a linear mapping due to the existence of the activation function.}
is sufficient to serve as an optimal reweighting function to align not only the correlation but also the joint distribution between variables. Mathematically, this finding could be formalized as Proposition~\ref{prop:gauss_linear_map}. All proofs in this section are postponed to Appendix \ref{appdix:theory}.



\begin{restatable}[Gaussian Probability Alignment]{proposition}{GausMap}\label{prop:gauss_linear_map}
    
    Suppose source data $\mtx{X}_s\in\mathbb{R}^{D\times T}$ and target data $\mtx{X}_t\in\mathbb{R}^{D\times T}$ follow $\mathcal{N}_s(\bm{\mu}_s,\bm{\Sigma}_s)$ and $\mathcal{N}_t(\bm{\mu}_t,\bm{\Sigma}_t)$, respectively.
    There exists a reweighting matrix $\mtx{A}\in\mathbb{R}^{D\times D}$ and a bias vector $\mtx{b} \in \mathbb{R}^D$, such that the multivariate joint probability of the reweighted target domain perfectly aligns with that of the source domain, that is for every $i,j=1,2,..., D$
    \begin{equation*}
        \mathrm{Pr}
        \left(\mtx{X}_s[i],\mtx{X}_s[j]\right) = 
        \mathrm{Pr}
        \left(\mtx{Y}[i],\mtx{Y}[j]\right)
    \end{equation*}
    where $\mtx{Y} = \mtx{A} \mtx{X}_t + \mtx{b}$ and $\mtx{b} = \mtx{0}$ for most MTS data.
\end{restatable}


This insight offers a promising direction for designing effective solutions without resorting to complex and computationally expensive methods. Importantly, even when the distributions of random variables are complex and difficult to characterize precisely, this simple linear mapping approach can still perfectly match the correlation between variables, thereby effectively mitigating correlation shifts, as shown in the following proposition. 

\begin{restatable}[Correlation Alignment]{proposition}{GeneralMap}\label{prop:gen_linear_map}
    Suppose source data $\mtx{X}_s\in\mathbb{R}^{D\times T}$ and target data $\mtx{X}_t\in\mathbb{R}^{D\times T}$ follow the source distribution $\mathcal{P}_s$ and the target distribution $\mathcal{P}_t$, respectively.
    There exists a reweighting matrix $\mtx{A}\in\mathbb{R}^{D\times D}$, such that the correlation of the source distribution and target distribution can be perfectly aligned, formally expressed as:
    \begin{align}
        &\mathrm{Corr}(\mathbf X_s) = \mathrm{Corr}(\mathbf A\mathbf X_t)
    \end{align}
\end{restatable}

\vspace{-1mm}
Intuitively, the interaction between variables can be considered as a fully-connected graph, where each node represents a variable and edges model the inter-variable dependency, and the reweighting matrix $\mtx{A}$ serves as the adjacency matrix of the graph.
However, in practical scenarios, solving the reweighting matrix $\mtx{A}$ is often computationally expensive and non-trivial, particularly when the distributions of the variables are highly complex. To address this, we leverage a Graph Attention Network (GAT) \cite{velickovic2017graph} to adaptively approximate the matrix $\textbf{A}$.
Mathematically, we formalize this insight through the following theorem:
\begin{restatable}[Attention Approximation]{theorem}{GATMap}\label{thm:gat}
    The optimal reweighting matrix $\mtx{A}$ in Proposition~\ref{prop:gen_linear_map} can be approximated by an attention matrix $\mtx{\Tilde{A}}$ with an arbitrary precision, where $\mtx{\Tilde{A}}$ is generated by a one-layer Graph Attention Network with an infinite hidden dimension.
\end{restatable}

\vspace{-1mm}
By integrating the GAT and TDCs, we finally propose \method{} to both well capture temporal dependencies and solve correlation shift.
Mathematically, the \method{} layer $\mtdfunc(\cdot)$ could be expressed as:
\begin{equation} \label{eq:formula_method}
    \begin{aligned}
        \mtdfunc(\mtx{X}) = &\mtx{X} + \operatorname{TDC}_{\uparrow} \left(\sigma(\operatorname{GAT}\left(\operatorname{TDC}_{\downarrow}\left(\mtx{X}\right)\right))\right)
    \end{aligned}
\end{equation}
where $\operatorname{TDC}_{\downarrow}$ and $\operatorname{TDC}_{\uparrow}$ represent the temporal convolution layers for down-project and up-project, respectively. Here, $\operatorname{GAT}$ represents one GAT layer on a fully connected graph. We prove in Appendix \ref{appdix:theorem2} that 
\method{} in Eq. \eqref{eq:formula_method} could also approximate the reweighting matrix $\mtx{A}$ in a manner similar to a one-layer GAT.


Similar to typical domain adapters in Eq. \eqref{eq:typ_adp}, we integrate \method{} within two consecutive blocks of a Transformer-based model. Since Transformers always consist of multiple blocks, this allows each instance of \method{} to make minor adjustments to the target domain's distribution. Cumulatively, these incremental adjustments are capable of mitigating large domain shift from the final block. The idea of gradually reducing domain shift is conceptually similar to but bears subtle difference from gradual domain adaptation \cite{he2024gradual}. Gradual domain adaptation leverages intermediate domains for supervision, which are often predefined. In contrast, \method{} does not rely on any intermediate domain, making it a more flexible and efficient solution for UDA.

\vspace{-1mm}
\subsection{Training Procedures} \label{sec:train}

This section introduces the training objective for \method{} on the unlabeled target domain. 
The goal is to preserve the pre-trained model's rich knowledge for accurate time series classification while leveraging \method{} to effectively minimize distribution shifts across domains.

To achieve this, we first propose the correlation alignment loss, specifically designed to address correlation shift. Given \method{}'s ability to adaptively reweight multivariate correlation, as discussed in Section \ref{sec:gat}, our objective is to align the correlation distribution of the source domain with that of the target domain transformed by \method{}.
Specifically, given the output of the $k$-th block $\mtx{H}^{(k)}_s$ and $\mtx{H}^{(k)}_t$ from the source domain and the target domain respectively, we minimize the Maximum Mean Discrepancy (MMD) \cite{gretton2012kernel} of the correlation distribution between $\mtx{H}^{(k)}_s$ and $\mtdfunc(\mtx{H}^{(k)}_t)$. Mathematically, the correlation alignment loss can be expressed as:
\begin{equation}
\vspace{-4mm}
\begin{small}
    \begin{aligned}
        \mathcal{L}_\mathtt{corr} = \sum_{k=1}^K \sum_{\substack{\mtx{H}^{(k)}_s \\ \mtx{H}^{(k)}_t}}\operatorname{MMD}\left(\operatorname{corr}\left(\mtx{H}^{(k)}_s\right), \operatorname{corr}\left(\mtdfunc\left(\mtx{H}^{(k)}_t\right)\right)\right) \nonumber
    \end{aligned}
\end{small}
\end{equation}
\noindent where $\operatorname{corr}(\mtx{H}) = \operatorname{vec} \left(\frac{\mtx{H}\mtx{H}^T}{\Vert \mtx{H} \Vert^2_\mathcal{F}}\right)$, $\operatorname{MMD}(\cdot, \cdot)$ denotes the MMD loss, and $\operatorname{vec}(\cdot)$ denotes the vectorization operator. 

Compared to directly aligning hidden features using MMD, the correlation alignment loss offers a unique advantage in terms of optimization difficulties. This is primarily because MMD assumes that data distributions are i.i.d., while time series data inherently exhibit non-i.i.d. characteristics. Consequently, directly applying MMD to align feature distributions often increases the difficulty of optimization, potentially leading to suboptimal performance. In contrast, correlation alignment loss focuses on aligning correlations rather than directly aligning raw features. Within the same domain, these correlations across variables tend to be more stable compared to the feature distributions. For instance, in a financial time series dataset, the correlation between stock prices of two closely related companies might remain consistent over time, even though the individual stock price values fluctuate significantly \cite{kim2016cross}. Thus, within a single domain, if we consider the correlation of MTS data as a new ``feature", this ``feature" tends to exhibit higher similarity across different samples. Consequently, using MMD to align correlations becomes less challenging in terms of optimization. Therefore,
this property makes correlation alignment loss a more stable and effective approach for reducing distributional discrepancies, particularly in MTS tasks, where temporal dependencies and multivariate correlation play a crucial role

In addition to mitigating correlation shift, it is crucial to enhance \method{}'s ability to understand the task and data accurately. To ensure that the features extracted by \method{} are beneficial for the classification task, we compute a classification loss $\mathcal{L}_\mathtt{c}$ on the labeled source domain. Furthermore, to improve \method{}'s understanding of the input features from the unlabeled target domain, we introduce a forecasting loss $\mathcal{L}_\mathtt{f}$ on the target domain, encouraging \method{} to accurately capture local temporal patterns. These two losses are discussed in Appendix \ref{appdix:train_obj} in detail. To sum up, the final loss can be computed as follows:
\begin{equation} \label{eq:final_loss_func}
    \mathcal{L} = \mathcal{L}_\mathtt{c} + \lambda_{\mathtt{corr}}\mathcal{L}_\mathtt{corr} + \lambda_\mathtt{f} \mathcal{L}_\mathtt{f}
\vspace{-1mm}
\end{equation}
where $\lambda_\mathtt{f}$ and $\lambda_\mathtt{corr}$ are two hyperparameters.
\vspace{-2mm}
\section{Experiments}

\begin{table*}[t]
    \centering
    \caption{Main accuracy results for MTS classification on the UDA task. The higher the accuracy is, the better. For three Transformer variants, the columns of `w/o \method', `w/ \method' and `$\Delta$' represent the accuracy without \method, the accuracy with \method, and the accuracy improvement due to \method. \textbf{Bold font} indicates the best performance across \textbf{all the methods}, and \underline{underline symbol} represents the best performance among \underline{UDA baselines}. 
    }
    \resizebox{\linewidth}{!}{
    \begin{tabular}{l|ccccc | cac | cac | cac}
        \toprule
        \textbf{Dataset} &   \multicolumn{5}{c|}{\textbf{UDA Baseline}} & \multicolumn{3}{c}{\textbf{Transformer}} & \multicolumn{3}{c}{\textbf{TimesNet}} & \multicolumn{3}{c}{\textbf{iTransformer}} \\
        \mapping{Source}{Target} & CORAL & Raincoat & CLUDA & SASA & UDApter & w/o \method & w/ \method & $\Delta$ & w/o \method & w/ \method & $\Delta$ & w/o \method & w/ \method & $\Delta$  \\
        
        \midrule
            HAR \mapping{24}{27} & 78.76 & \underline{96.88} & 82.14 & 86.72 & 96.46 & 91.81 & {98.23} & 6.42 & 93.69 & 97.34 & 3.65 & 82.30 & \textbf{99.11} & 16.81\\
            HAR \mapping{3}{13} & 63.63 & \underline{91.67} & 77.55 & 78.78 & 90.90 & 79.79 & \textbf{98.98} & 19.19 & 84.96 & 87.86 & 2.90 & 75.75 & {97.97} & 22.22\\
            HAR \mapping{16}{13} & 47.47 & \underline{71.87} & 69.39 & 61.61 & 66.67 & 73.96 & 77.78 & 3.82 & 67.67 & {83.84} & 16.17 & 74.74 & \textbf{85.86} & 11.12\\
            HAR \mapping{3}{8} & 51.76 & 78.13 & \underline{78.57} & 64.70 & 71.76 & 54.11 & 75.12 & 21.01 & 64.70 & \textbf{92.92} & 28.22 & 74.11 & {91.77} & 17.66\\
            HAR \mapping{19}{2} & 61.53 & \underline{76.56} & 60.00 & 69.23 & 59.34 & 53.84 & 73.52 & 19.68 & 53.84 & {82.41} & 28.57 & 59.34 & \textbf{84.61} & 25.27\\
            HAR \mapping{11}{28} & 60.86 & 73.95 & 64.91 & \underline{76.52} & 66.95 & 66.95 & {77.40} & 10.45 & 70.43 & \textbf{80.00} & 9.57 & 57.39 & {77.40} & 20.01\\
            HAR \mapping{16}{10} & 50.56 & \underline{71.88} & 68.42 & 56.17 & 61.79 & 35.95 & 55.05 & 19.10 & 62.92 & {72.91} & 9.99 & 67.41 & \textbf{87.64} & 20.23\\
            HAR \mapping{25}{10} & 19.10 & 57.81 & \underline{57.89} & 56.18 & 56.18 & 48.31 & 57.40 & 9.09 & 46.06 & \textbf{65.17} & 19.11 & 47.19 & \textbf{65.17} & 17.98\\
            HAR \mapping{18}{10} & 37.07 & 48.43 & \underline{57.89} & 37.07 & 46.06 & 35.95 & {59.55} & 23.60 & 38.20 & \textbf{69.66} & 31.46 & 40.44 & 48.51 & 8.07\\
            HAR \mapping{19}{10} & 44.94 & \underline{50.21} & 49.12 & 39.32 & 43.82 & 37.07 & 49.48 & 12.41 & 42.94 & 46.56 & 3.62 & 37.07 & \textbf{50.56} & 13.49\\
            \hline
            HAR Average & 51.57 & \underline{71.74} & 66.59 & 62.63 & 65.99 & 57.77 & 72.24 & 13.87 & 62.54 & {77.87} & 15.33 & 61.57 & \textbf{78.86} & 17.29\\
        \midrule
            WISDM \mapping{12}{9} & 82.71 & \underline{91.35} & 82.50 & 75.30 & 83.95 & 82.50 & 85.19 & 2.69 & 72.83 & \textbf{92.60} & 19.77 & 66.67 & 90.12 & 23.45\\
            WISDM \mapping{5}{31} & 59.03 & 80.72 & \underline{82.93} & 75.90 & \underline{82.93} & 75.90 & 74.70 & -1.20 & 81.92 & 81.92 & 0.00 & 65.06 & \textbf{83.13} & 18.07\\
            WISDM \mapping{25}{31} & 48.19 & \underline{61.44} & 53.66 & 43.47 & 59.03 & 56.62 & \textbf{61.44} & 4.82 & 57.83 & 60.24 & 2.41 & 44.57 & 59.03 & 14.46\\
            WISDM \mapping{0}{30} & \underline{\textbf{65.04}} & 61.16 & 62.75 & 63.10 & 59.03 & 58.22 & 60.19 & 1.97 & 58.22 & 61.16 & 2.94 & 58.25 & 62.14 & 3.89\\
            WISDM \mapping{10}{22} & 61.67 & 73.33 & \underline{76.67} & 51.66 & 71.67 & 71.67 & 76.67 & 5.00 & 73.00 & 76.67 & 3.67 & 56.67 & \textbf{78.33} & 21.66\\
            WISDM \mapping{12}{2} & 36.59 & 53.65 & \underline{63.41} & 58.53 & 41.46 & 48.78 & 62.19 & 13.41 & 48.78 & 46.34 & -2.44 & 51.21 & \textbf{67.07} & 15.86\\
            WISDM \mapping{6}{11} & 43.42 & \underline{56.57} & 56.10 & 47.36 & 41.46 & 43.36 & 42.10 & -1.26 & 56.57 & 59.21 & 2.64 & 27.63 & \textbf{63.15} & 35.52\\
            WISDM \mapping{11}{21} & 28.84 & 38.46 & \underline{58.54} & 40.38 & 30.76 & 18.84 & 19.23 & 0.39 & 38.46 & \textbf{59.61} & 21.15 & 17.30 & 55.76 & 38.46\\
            WISDM \mapping{19}{3} & 7.69 & 15.38 & \underline{\textbf{51.22}} & 50.00 & 23.07 & 19.23 & 38.46 & 19.23 & 23.07 & 42.30 & 19.23 & 19.92 & 19.23 & -0.69\\
            WISDM \mapping{3}{11} & 38.16 & 21.36 & \underline{48.78} & 25.00 & 15.78 & 17.10 & 18.42 & 1.32 & 15.78 & \textbf{60.52} & 44.74 & 13.15 & 18.42 & 5.27\\
            \hline
            WISDM Average & 47.13 & 55.34 & \underline{63.66} & 53.07 & 50.91 & 49.22 & 53.86 & 4.64 & 52.65 & \textbf{64.06} & 11.41 & 42.04 & 59.64 & 17.60\\
        \midrule
            HHAR \mapping{7}{3} & 55.57 & \underline{94.08} & 85.09 & 79.86 & 86.87 & 61.26 & 88.96 & 27.70 & 83.58 & \textbf{94.96} & 11.38 & 84.87 & 92.24 & 7.37\\
            HHAR \mapping{6}{7} & 56.99 & \underline{84.37} & 76.15 & 58.24 & 83.50 & 74.15 & 84.55 & 10.40 & 58.87 & 89.56 & 30.69 & 75.15 & \textbf{93.32} & 18.17\\
            HHAR \mapping{6}{3} & 48.14 & \underline{74.33} & 65.79 & 66.52 & 65.86 & 57.55 & 83.58 & 26.03 & 62.36 & 75.27 & 12.91 & 67.36 & \textbf{84.47} & 17.11\\
            HHAR \mapping{6}{5} & 45.47 & \underline{75.58} & 45.47 & 61.70 & 45.64 & 47.38 & 66.54 & 19.16 & 49.90 & 70.98 & 21.08 & 42.15 & \textbf{76.40} & 34.25\\
            HHAR \mapping{7}{5} & 36.75 & \underline{63.47} & 48.06 & 57.05 & 45.64 & 43.52 & 63.63 & 20.11 & 54.35 & 66.53 & 12.18 & 43.32 & \textbf{70.99} & 27.67\\
            HHAR \mapping{0}{7} & 41.97 & \underline{68.32} & 33.89 & 34.34 & 62.83 & 44.25 & 71.81 & 27.56 & 38.20 & \textbf{68.47} & 30.27 & 56.57 & 66.97 & 10.40\\
            HHAR \mapping{4}{0} & 22.54 & 23.66 & \underline{34.73} & 25.16 & 25.16 & 23.72 & 24.94 & 1.22 & 23.63 & 28.67 & 5.04 & 27.32 & \textbf{31.29} & 3.97\\
            HHAR \mapping{3}{0} & 26.70 & 17.41 & \underline{\textbf{35.15}} & 22.10 & 22.10 & 9.63 & 23.20 & 13.57 & 17.25 & 20.56 & 3.31 & 25.16 & 29.54 & 4.38\\
            HHAR \mapping{2}{7} & 5.42 & \underline{54.68} & 26.36 & 32.98 & 41.33 & 34.65 & 58.89 & 24.24 & 41.54 & \textbf{69.10} & 27.56 & 44.25 & 63.63 & 19.38\\
            HHAR \mapping{1}{0} & 36.19 & \underline{57.32} & 47.65 & 45.90 & 44.30 & 41.27 & 58.69 & 17.42 & 44.30 & 60.97 & 16.67 & 48.95 & \textbf{64.28} & 15.33\\
            \hline
            HHAR Average & 37.57 & \underline{61.44} & 49.83 & 48.39 & 52.32 & 43.74 & 62.48 & 18.74 & 47.40 & {64.51} & 17.11 & 51.51 & \textbf{67.31} & 15.80\\
            \midrule
            Boiler \mapping{1}{2} & 93.76 & 97.05 & 97.29 & \underline{97.33} & 92.64 & 91.98 & 97.86 & 5.88 & 97.86 & \textbf{98.15} & 0.29 & 91.51 & \textbf{98.15} & 6.64\\
            Boiler \mapping{3}{2} & 87.16 & 95.02 & 87.16 & \underline{96.05} & 92.17 & 90.83 & \textbf{98.15} & 7.32 & 92.17 & 97.84 & 5.67 & 97.47 & \textbf{98.15} & 0.68\\
            \hline
            Boiler Average & 90.46 & 96.03 & 92.22 & \underline{96.69} & 92.41 & 91.41 & 98.00 & 6.59 & 95.02 & 98.00 & 2.98 & 94.49 & \textbf{98.15} & 3.66\\
        \bottomrule
    \end{tabular}
    }
    \label{tab:comprehensive_compare}
    \label{tab:my_label}
\end{table*}

\subsection{Experimental Settings}
\textbf{Datasets.}
We conduct experiments on 4 real-world datasets, including HAR \cite{anguita2013public}, WISDM \cite{wisdm_smartphone_and_smartwatch_activity_and_biometrics_dataset__507}, HHAR \cite{stisen2015smart}, and Boiler \cite{cai2021time}. For HAR, HHAR, and WISDM datasets, we rank all possible source-target domain pairs based on the magnitude of domain shift, dividing them into 10 groups in the ascending order. From each group, we select one source-target domain pair for evaluation. For Boiler, due to its limited number of domains, we choose the domain pairs with the largest or the smallest domain shift. Detailed explanations of datasets and domain pair selection are provided in Appendix \ref{appdix:dataset} and \ref{appdix:domain_select}, respectively.

\textbf{Baselines.} We compare \method{} with three different types of UDA methods, including (1) correlation-related UDA: CORAL, (2) MTS UDA: SASA, CLUDA, and Raincoat, and (3) adapter-based UDA: UDApter. Descriptions of baseline methods are in Appendix \ref{appdix:baseline}

\textbf{Parameter settings.} Unless otherwise specified, we use default hyperparameter settings in the released code of corresponding publications. For \method, we use TDCs with a kernel size $r=5$ and a padding of 2. For training, we use Adam optimizer with a learning rate of 1e-4, and set $\lambda_\mathtt{c}=0.5$ and $\lambda_\mathtt{f}=0.5$. We evaluate the performance of \method{} on three different Transformer-based MTS models: Transformer \cite{vaswani2017attention}, TimesNet \cite{wu2022timesnet}, and iTransformer \cite{liu2023itransformer}. The implementation details are provided in Appendix \ref{appdix:implemetation}.

\subsection{Experimental Results}
\textbf{Main results.} The main evaluation results of the accuracy are presented in Table \ref{tab:comprehensive_compare}. On each dataset in the table, the difficulty of UDA tasks for source-target domain pairs increases progressively from top to bottom. 
The experimental results reveal two noteworthy conclusions: \textbf{(1) \method{} significantly enhances the UDA classification performance of Transformer-based MTS models.} Specifically, on four datasets \method{} improves the average accuracy of Transformer, TimesNet, and iTransformer on the target domain by $17.29\%$, $17.60\%$, $15.80\%$, and $3.66\%$ accucary, respectively. Even under scenarios with the largest domain shifts, such as HAR \mapping{19}{10} and HHAR \mapping{1}{10}, \method{} demonstrates robust performance, delivering $9.84\%$ and $16.47\%$ improvement on average for all three models. These results strongly validate the effectiveness of \method{}, even in scenarios with large domain shifts. \textbf{(2) \method{}-enhanced MTS models outperform state-of-the-art (SOTA) baselines in classification accuracy.} Across all four datasets, \method{}-enhanced models achieve the best performance, with average accuracy improvements of $7.12\%$, $0.40\%$, $5.87\%$, and $1.46\%$ accuracy, respectively, compared to SOTA baselines. These results highlight the superiority of \method{} in addressing UDA challenges for multivariate time series data.

\begin{figure*}
    \centering
    \includegraphics[width=\linewidth]{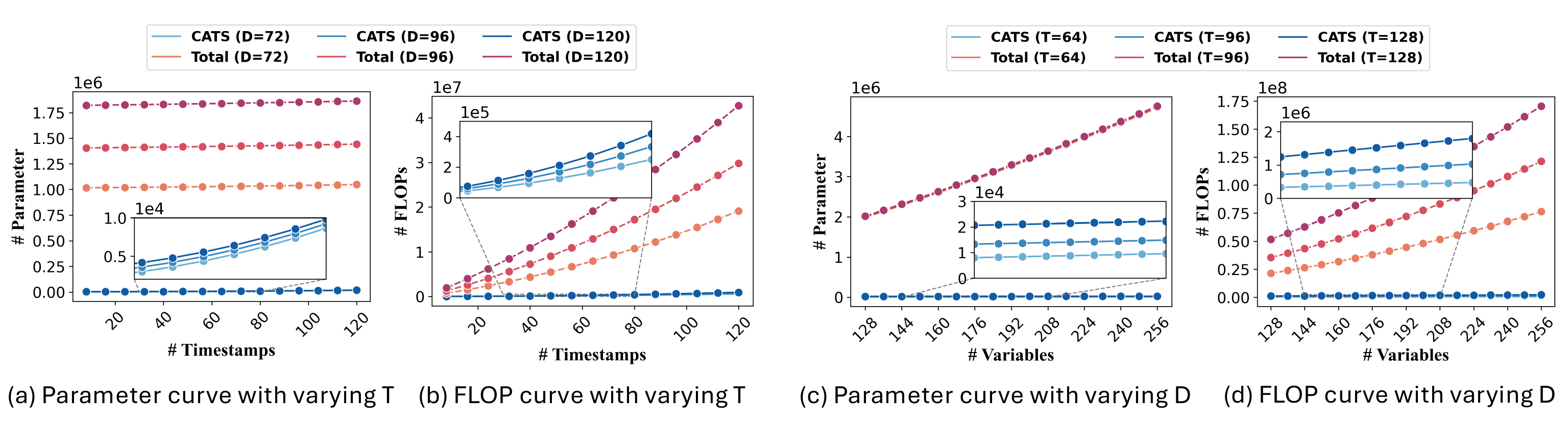}
    \vspace{-8mm}
    \caption{The parameter (FLOP) curve of \method{} on Transformer with varying variable number $D$ and time length $T$. The three red curves represent the parameter counts (or FLOPs) of the full Transformer model, while the three blue curves represent \method{} alone. In Figure (c), the three red curves overlap due to their relatively small differences compared to the large overall values.}
    \label{fig:scalable_exp}
\end{figure*}
 
\textbf{Scalability evaluation.} To validate the scalability of \method{}, we adjust the time series length $T$ and the number of variables $D$ of the Transformer, recording the parameter count and FLOPs (Floating Point Operations per Second) for \method{} and the full model. The experimental results are shown in Figure \ref{fig:scalable_exp}. The results reveal that, regardless of the values of $T$ and $D$, \method{} consistently requires two orders of magnitude fewer parameters and FLOPs compared to the full model. Interestingly, as the hidden layer dimension $D$ increases, the parameter count and FLOPs of the Transformer exhibit quadratic growth, whereas \method{} scales linearly. This observation confirms \method{}'s suitability for large-scale MTS tasks with varying input dimensions.

\begin{figure}
    \centering
    \includegraphics[width=\linewidth]{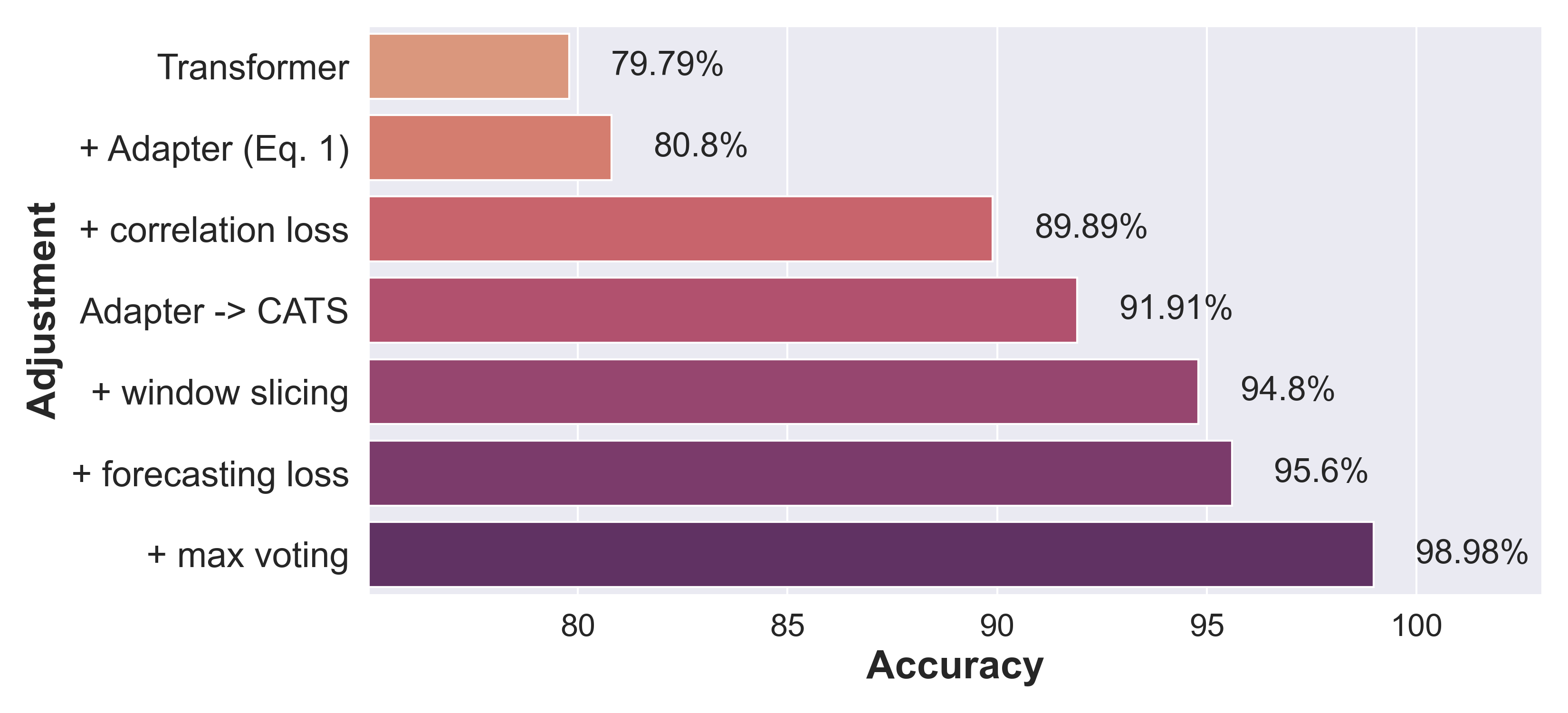}
    \vspace{-8mm}
    \caption{Step-by-Step accuracy improvement on HAR dataset from vanilla Transformer to Transformer enhanced by \method{}.}
    \label{fig:ablation}
    \vspace{-3mm}
\end{figure}

\textbf{Ablation study.} To validate the effectiveness of \method{} and the proposed loss function in Eq. \eqref{eq:final_loss_func}, we conducted stepwise ablation experiments on the HAR \mapping{3}{13} scenario, using a Transformer as the backbone model. Incremental adjustments, from the vanilla Transformer to the \method{}-enhanced Transformer, are detailed in Appendix \ref{appdix:incremental_adjust}. The results, presented in Figure \ref{fig:ablation}, demonstrate that every proposed improvement in this paper contributes positively to accuracy. Notably, introducing the correlation alignment loss (third row in Figure \ref{fig:ablation}) and replacing the typical adapter in Eq. \eqref{eq:typ_adp} with \method{} (fourth row) lead to significant enhancements in target domain performance. These findings highlight the effectiveness of \method{} and the proposed loss function in mitigating domain shift and improving target domain classification accuracy.

\vspace{-2mm}
\section{Related Works} 
\vspace{-2mm}
\textbf{Unsupervised domain adaptation.} Unsupervised domain adaptation (UDA), which utilizes labeled data from a source domain to predict labels in an unlabeled target domain, is a widely research area across various application fields \cite{ganin2015unsupervised, zhang2018task, ramponi2020neural, liu2021cycle}. Existing DA methods can broadly be grouped into three categories. (1) \textit{Adversarial training approaches} use a domain discriminator to differentiate between source and target domains while training a classifier to extract domain-invariant features \cite{hoffman2018cycada, long2018conditional, tzeng2015simultaneous}. (2) \textit{Multi-task supervision approaches} introduce auxiliary self-supervised tasks, such as data augmentation \cite{volpi2018adversarial} or reconstruction \cite{ghifary2016deep, zhuo2017deep}, to enhance feature learning in the target domain. (3) \textit{Statistical divergence approaches} minimize distributional discrepancies between domains using metrics such as maximum mean discrepancy (MMD) \cite{yan2017mind, zhang2020discriminative, yan2019weighted}, optimal transport distance \cite{courty2017joint, courty2016optimal}, and contrastive domain discrepancy (CDD) \cite{kang2019contrastive}. CORAL-based methods \cite{sun2016deep, lee2019coral+, li2022coral++} also focus on aligning correlations. However, unlike \method{}, they fail to model the temporal dependencies of MTS data and overlook the importance of aligning the means of source and target distributions, as discussed in Appendix \ref{appdix:compare_coral}.

\vspace{-2mm}
\paragraph{Unsupervised domain adaptation for time series.} While adaptation methods have achieved significant success in computer vision, relatively fewer approaches have been developed to address the unique challenges of domain adaptation for time series data. CoDATS \cite{wilson2020multi} employ domain discriminators for temporal feature alignment. SASA \cite{cai2021time} aligns invariant unweighted spare associative structures for time series data. RainCoat \cite{he2023domain} utilizes MMD to minimize frequency feature distribution in a polar coordinate across domains. Additionally, CLUDA \cite{ozyurt2022contrastive} leverage contrastive learning to enhance model robustness with data augmentations, while LogoRA \cite{zhang2024logora} combines global and local feature analysis to maintain domain-invariant representations for complex time series structures.

\vspace{-2mm}
\paragraph{Graph Neural Networks.} GNNs are effective for capturing dependencies within graphs. Graph Convolutional Networks (GCNs) \cite{zhang2019graph, kipf2016semi} aggregate neighbor information by utilizing a localized first-order approximation of spectral graph convolutions. Graph Attention Networks (GATs) \cite{velivckovic2017graph} implement attention mechanisms that dynamically weigh the contributions of neighboring nodes. GRAND \cite{feng2020graph} learns node representations by randomly dropping nodes to augment data and enforcing the consistency of predictions among augmented data. GraphSAGE \cite{hamilton2017inductive} generates embeddings for unseen nodes by sampling and aggregating features from the local neighborhood. For more recent works on GNNs, please see \cite{sharma2024survey, ju2024survey, khoshraftar2024survey, shao2024distributed}. 

\section{Conclusion}
In this paper, we study the problem of unsupervised domain adaptation for multivariate time series classification.  We begin by identifying a previously overlooked domain shift in MTS data: correlation shift, where correlations between variables vary across domains.
To mitigate this shift, we propose a scalable and parameter-efficient adapter, \method{}, serving as a plug-and-play technique compatible with various Transformer variants. Supported by a solid theoretical foundation for mitigating correlation shift, \method{} effectively captures dynamic temporal patterns while adaptively reweighting multivariate correlations. To further reduce correlation discrepancies, we introduce a correlation alignment loss, which aligns multivariant correlations across domains, addressing the non-i.i.d. nature of MTS data.
Extensive evaluations on real-world datasets demonstrate that \method{} significantly improves the accuracy of Transformer backbones while introducing minimal additional parameters.

\section*{Impact Statement}
Our work focuses solely on the technical challenge of domain adaptation for multivariate     time series classification and does not involve any elements that could pose ethical risks.

\bibliographystyle{ACM-Reference-Format}
\bibliography{contents/ref}

\newpage
\appendix

\onecolumn
\textbf{\Large Appendix}

\addtocontents{toc}{\protect\setcounter{tocdepth}{2}}

\tableofcontents

\clearpage

\section{Empirical Evidence of Correlation Shift}\label{appdix:emp_cor_shift}
In this section, we conduct an empirical analysis on the Human Activity Recognition (HAR) dataset \cite{anguita2013public} to demonstrate the prevalence of correlation shift. The HAR dataset consists of 30 domains. Then we iterate through the 30 domains in HAR, treating each domain as the source domain while considering the remaining 29 domains as target domains. Our objective is to examine whether multivariate correlations differ significantly between samples from the source and target domains.

For each sample from either the source or the target domain, we compute the multivariate correlation matrix, which is an $D \times D$ high-dimensional structure with $D$ being the number of variables. However, directly analyzing such high-dimensional matrices is challenging. Hence, we calculate the element-wise mean of the correlation matrices for each domain. Mathematically, if the mean values for the correlation matrices from the source and target domains come from different distributions, it implies a significant difference in the distribution of the overall correlations.

To formally test this, we set the null hypothesis $\mathcal{H}_0$ : \textit{the mean distributions of the source and target domains originate from the same underlying distribution}. We then apply the Mann-Whitney U test \cite{mcknight2010mann} to verify $\mathcal{H}_0$. If the p-value is less than 0.05, it indicates a statistical confidence of over $95\%$ in rejecting $\mathcal{H}_0$. This rejection implies that the mean values are from different distributions, confirming the statistical significance of the correlation shift between the source and target domains. 

Finally, for every source domain, we calculate the rate of target domains with a significant correlation shift among the rest 29 domains.
The experiment result is provided in Figure \ref{fig:p_value_x_shift}. The {\em x}-axis represents the source domain ID, and the {\em y}-axis values of the orange bars indicate the rate of target domains with significant correlation shifts. The red dashed line in Figure \ref{fig:p_value_x_shift} marks the average rate of correlation shift, which is 78\%. These findings provide clear evidence that correlation shifts are prevalent in multivariate time series datasets, highlighting the need to address such shifts.

\section{Empirical Comparison Between TDC and Linear Matrices}\label{appdix:compare_tdc_linear}
To assess the representative learning ability between TDC and linear matrices in adapters, we compare the performance of these two adpaters on HAR dataset with the domain 1 being the source domain and domain 10 being the target domain. Furthermore, we leverege TimesNet as the backbone, and set both the time length $T$ and the hidden dimension $D$ as 128 to make their parameters compatible. Then after pretraining the backbone on the source domain, we only train these adapters on the target domain with accessible labels. We use Adam optimizer with a learning rate of 1e-4 during training.

\section{Theoretical Analysis of Correlation Shift} \label{appdix:theory}


\subsection{Proof of Correlation Alignment}\label{appdix:prop2}
\GeneralMap*
    
\begin{proof}
    Let $\bm{\Sigma}_t = \mathbb{E}_{\mtx{X}_t \sim \mathcal{P}_t}\left[ \left(\mtx{X}_t - \mathbb{E}\mtx{X}_t\right)\left(\mtx{X}_t - \mathbb{E}\mtx{X}_t\right)^T \right]$ and $\hat{\bm{\Sigma}}_t = \mathbb{E}_{\mtx{Y}}\left[ \left(\mtx{Y} - \mathbb{E}\mtx{Y}\right)\left(\mtx{Y} - \mathbb{E}\mtx{Y}\right)^T \right]$ where $\mtx{Y} = \mtx{A} \mtx{X}_t$. Then based on the spectral theorem, we can decompose the covariance matrices $\bm{\Sigma}_s$ and $\bm{\Sigma}_t$:
    \begin{align}
        \bm{\Sigma}_s = \mtx{U}_s \bm{\Lambda}_s \mtx{U}_s^T \\
        \bm{\Sigma}_t = \mtx{U}_t \bm{\Lambda}_t \mtx{U}_t^T
    \end{align}
    Then since $\mtx{Y} = \mtx{A} \mtx{X}$, the covariance matrix $\hat{\bm{\Sigma}}_t$ could be expressed as 
    \begin{equation}
        \begin{aligned}
            \hat{\bm{\Sigma}}_t &= \mtx{A} \bm{\Sigma}_t \mtx{A}^T \\
            &= \mtx{A} \mtx{U}_t \bm{\Lambda}_t \mtx{U}_t^T \mtx{A}^T
        \end{aligned}
    \end{equation}
        Let $\mtx{A} = \mtx{U}_s\Lambda_s^{\frac{1}{2}}\bm{\Lambda}_t^{-\frac{1}{2}} \mtx{U}_t^T$. Then we have 
    \begin{equation}
        \begin{aligned}
            \hat{\bm{\Sigma}}_t &= \mtx{A} \mtx{U}_t \bm{\Lambda}_t \mtx{U}_t^T \mtx{A}^T \\
            &= \mtx{U}_s\bm{\Lambda}_s^{\frac{1}{2}}\bm{\Lambda}_t^{-\frac{1}{2}} \mtx{U}_t^T \mtx{U}_t \bm{\Lambda}_t \mtx{U}_t^T \mtx{U}_t \bm{\Lambda}_t^{-\frac{1}{2}} \bm{\Lambda}_s^{\frac{1}{2}} \mtx{U}_s^T \\
            &= \mtx{U}_s \bm{\Lambda}_s \mtx{U}_s^T \\
            &= \bm{\Sigma_}s
        \end{aligned}
    \end{equation}
    Therefore, with the reweighting matrix, the covariance matrix of $\mtx{Y}$ on the target domain could equal to the covariance matrix of $\mtx{X}_s$ on the source domain. Then, since the correlation matrix is only defined by the covariance matrix as shown in Eq. \eqref{eq:corr_dfn}, the correlation matrices for $\mtx{Y}$ on the target domain and $\mtx{X}_s$ on the source domain are totally the same.
\end{proof}


\subsection{Proof of Gaussian Probability Alignment.} \label{appdix:prop1}

\GausMap*
\begin{proof}
    Based on Proposition \ref{prop:gen_linear_map}, it is obvious that $\bm{\Sigma}_s = \bm{\hat{\Sigma}}_t$ where $\bm{\hat{\Sigma}}_t$ is the covariance matrix of $\mtx{Y}$. Then the mean of $\mtx{Y}$ could be expressed as 
    \begin{equation}
        \begin{aligned}
            \mathbb{E} [\mtx{Y}] &= \mtx{A} \mathbb{E} [\mtx{X}] + \mtx{b
            } 
        \end{aligned}
    \end{equation}
    Therefore, as long as $\mtx{b} = (\mtx{I} - \mtx{A}) \mathbb{E}\mtx{X}$, we will have $\mathbb{E}\mtx{Y} = \mathbb{E} \mtx{X}$. In the real world, it is quite common to normalize the MTS data during data preprocessing \cite{liu2023itransformer, wu2022timesnet, wang2024tssurvey}. Under this circumstance, the expectation of normalized data would be zero, and hence $\mtx{b} = \mtx{0}$.

    Finally, since the covariances and means of two Gaussian distribution is the same, then these two distribution are also the same. Therefore, we have $\mathrm{Pr}(\mtx{X}_s[i], \mtx{X}_s[j]) = \mathrm{Pr}(\mtx{Y}[i], \mtx{Y}[j])$.
\end{proof}


\subsection{Proof of Attention Approximation} \label{appdix:theorem1}

\GATMap*

\begin{proof}
    Our proof proceeds as follows: First, we leverage the universal approximation theorem to establish that an infinitely wide Multi-Layer Perceptron (MLP) can effectively approximate the reweighting matrix $\mtx{A}$. Then, we demonstrate that a one-layer GAT possesses the same learning capacity as an infinite-width MLP under appropriate conditions. This equivalence allows us to conclude that a one-layer GAT can effectively learn and approximate the matrix $\mtx{A}$, thereby completing the proof. 

    \textbf{Step 1. Revisit the Universal Approximation Theorem. } First, the Universal Approximation Theorem \cite{leshno1993multilayer, hornik1991approximation, hornik1989multilayer} states: 
    
    \textit{Let $f: \mathbb{R}^n \rightarrow \mathbb{R}$ be a continuous function defined on a compact subset, and let $\sigma(\cdot)$ be a nonlinear, measurable activation function. For any $\epsilon > 0$, there exists a MLP $F(t; \theta)$ with a single hidden layer of infinite width, such that:
    \begin{equation}
        \sup_{t\in \mathcal{T}} | f(t) - F(t; \theta) | < \epsilon
    \end{equation}
    where $\mathcal{T}$ is the compact input space, and $\theta$ represents the network parameters.
    }

    \textbf{Step2. Approximate the reweighting matrix by an MLP.}
    Now, let us consider a continuous, differentiable function $f: \mathbb{R} \rightarrow \mathbb{R}$ such that $f(i \cdot D + j ) = \mtx{A}[i, j]$, where $i, j$ are indices of the reweighting matrix $\mtx{A}$ and $D$ is the matrix dimension. Based on the Universal Approximation Theorem, we can approximate $f(\cdot)$ using an MLP with a single hidden layer of dimension $M$, when $M$ is quite large:
    \begin{equation}
        f(t) \approx F(t) = \mtx{w}_2^T \sigma(\mtx{w}_1 t) 
    \end{equation}
    where $\mtx{w}_1 \in \mathbb{R}^M$ and $\mtx{w}_1 \in \mathbb{R}^M$ are two learnable parameters. By stacking the approximated elements, we can have a reconstructed matrix $\mtx{A}_{mlp}[i, j] =  F(i \cdot D + j) \approx \mtx{A}[i, j]$. Therefore, the reweight output $\mtx{Y}_{mlp}$ in Proposition \ref{prop:gen_linear_map} could be expressed as:
    \begin{equation} \label{eq:final_mlp_formula}
        \begin{aligned}
            \mtx{Y}_{mlp}[i] &= \left(\mtx{A}_{mlp} \mtx{X} \right) [i] \\
            &= \sum_{d=1}^D \mtx{A}_{mlp}[i, d] \mtx{X}[d] \\
            &= \sum_{d=1}^D \mtx{w}_2^T \sigma(\mtx{w}_1 t_{i, d}) \mtx{X}[d] 
        \end{aligned}
    \end{equation}
    where $t_{i, d}$ is the the input for the row index $i$ and column index $d$.

    \textbf{Step3. Relating an MLP with a GAT.}
    Then we aim to show that a GAT layer has the same learning ability as an MLP. Let $\mtx{X}$ and  $\mtx{Y}_{gat}$ represent the input and output of the GAT layer, respectively.  To align with the notations used in the original GAT paper \cite{velivckovic2017graph}, we denote $\mtx{x}_i$ as the i-th row of $\mtx{X}$, i.e., $\mtx{x}_i = \mtx{X}[i]$. Then, The formula of a GAT layer could be expressed as follows:
    \begin{subequations}
        \begin{align}
            \mtx{Y}_{gat}[i] &= \sum_{j} \alpha_{i, j} \mtx{W} \mtx{x}_j, \\
            \alpha_{i, j} &= \frac{\exp\left( \operatorname{LeakyReLU} \left(\mtx{a}^T \left[ \mtx{W} \mtx{x}_i || \mtx{W} \mtx{x}_j \right] \right) \right)}{\sum_k \exp\left( \operatorname{LeakyReLU} \left( \mtx{a}^T \left[ \mtx{W} \mtx{x}_i || \mtx{W} \mtx{x}_k \right] \right)\right)} \label{subeq:gat_atten}
        \end{align}
    \end{subequations}
    where $\mtx{W} \in \mathbb{R}^{M \times M}$ and $\mtx{a} \in \mathbb{R}^{M}$ with $M$ being the input dimension. Here, $||$ represents the concatenation operation, and $||_k$ represents the concatenation operation over all the possible element with the subscripts $k$. 
    
    By defining a special nonlinear, measurable activation function $\Tilde{\sigma} ([\mtx{x}_i || \mtx{x}_j \Vert_{k\neq i,j } \mtx{x}_k]) = \frac{\exp\left( \operatorname{LeakyReLU} \left(\left[ \mtx{x}_i || \mtx{x}_j \right] \right) \right)}{\sum_k \exp\left( \operatorname{LeakyReLU} \left( \left[ \mtx{x}_i || \mtx{x}_k \right] \right)\right)}$, we could further simplified the attention coefficient into the following expression:
    \begin{equation}
        \begin{aligned}
            \alpha_{i, j} 
            &= \frac{\exp\left( \operatorname{LeakyReLU} \left(\mtx{a}^T \left[ \mtx{W} \mtx{x}_i || \mtx{W} \mtx{x}_j \right] \right) \right)}{\sum_k \exp\left( \operatorname{LeakyReLU} \left( \mtx{a}^T \left[ \mtx{W} \mtx{x}_i || \mtx{W} \mtx{x}_k \right] \right)\right)} \\
            &= \Tilde{\sigma} \left( \left[ \mtx{a}^T \mtx{W} \mtx{x}_i || \mtx{a}^T \mtx{W} \mtx{x}_j ||_{k \neq i, j} \mtx{a}^T \mtx{W} \mtx{x}_k\right] \right) \\
            &= \Tilde{\sigma} \left( \mtx{a}^T \mtx{W} \left[ \mtx{x}_i, \mtx{x}_j, \mtx{x}_1\dots, \mtx{x}_k, \dots, \mtx{x}_D \right] \right) \ (\text{where} \ k\neq i, j) \\
            &= \Tilde{\sigma}\left( \Tilde{\mtx{a}}^T \mtx{T}_{i, j} \right)
        \end{aligned}
    \end{equation}
    where $\left[\mtx{x}_i, \mtx{x}_j \right]$ represents the matrix stacked by the vector $\mtx{x}_i$ and $\mtx{x}_j$ along a new dimension. Here, $\Tilde{\mtx{a}} = \mtx{W}^T \mtx{a} \in \mathbb{R}^{M}$ and $\mtx{T}_{i,j } = \left[ \mtx{x}_i, \mtx{x}_j, \mtx{x}_1\dots, \mtx{x}_k, \dots, \mtx{x}_D \right] \in \mathbb{R}^{M \times D}$. Hence, the output $\mtx{Y}_{gat}$ could be simplified as:
    \begin{equation} \label{eq:final_gat_formula}
    \begin{aligned}
         \mtx{Y}_{gat}[i] &= \sum_{j=1}^D \Tilde{\sigma}\left( \Tilde{\mtx{a}}^T \mtx{T}_{i, j} \right) \mtx{W} \mtx{x}_j \\
         &=  \sum_{d=1}^D \Tilde{\sigma}\left( \Tilde{\mtx{a}}^T \mtx{T}_{i, d} \right) \mtx{W} \mtx{X}[d]
    \end{aligned}
    \end{equation}
   Obviously, Eq. \eqref{eq:final_gat_formula} shares a strong similarity with Eq. \eqref{eq:final_mlp_formula}. Intuitively, this similarity suggests that a GAT layer could exhibit learning behavior analogous to that of an MLP, making it capable of approximating the linear reweighting matrix $\mtx{A}$. It would be totally fine to stop here and conclude the proof, but we can still go a step further to rigorously show that the outputs of these two neural networks are element-wise equivalent.

   \textbf{Step 4. Element-wise comparison between the outputs.}
   First, let us expand the expression of $\mtx{Y}_{mlp}$ as follows:
   \begin{equation}\label{eq:mlp_final_expand}
       \begin{aligned}
           \mtx{Y}_{mlp} &= \sum_{d=1}^D \mtx{w}_2^T \sigma(\mtx{w}_1 t_{i, d}) \mtx{X}[d] \\
           &= \sum_{d=1}^D \sum_{m=1}^M \mtx{w}_2[m] \sigma\left(\mtx{w}_1 [m] t_{i,d}\right) \mtx{X}[d]
       \end{aligned}
   \end{equation}
   Similarly, we can also expand the expression of $\mtx{Y}_{gat}$ as follows:
   \begin{equation}\label{eq:gat_final_expand}
       \begin{aligned}
           \mtx{Y}_{gat} &= \sum_{d=1}^D \Tilde{\sigma}\left( \Tilde{\mtx{a}}^T \mtx{T}_{i, d} \right) \mtx{W} \mtx{X}[d] \\
           &= \sum_{d=1}^D \mtx{W} \Tilde{\sigma} \left( \sum_{m=1}^M \Tilde{\mtx{a}}[m] \mtx{T}_{i, j}[m] \right) \mtx{X}[d]
       \end{aligned}
   \end{equation}

   It is evident that the outputs of the GAT and MLP are essentially identical. The only difference lies in the order of summation and activation functions in Eq. \eqref{eq:mlp_final_expand} and Eq. \eqref{eq:gat_final_expand}. However, since the Universal Approximation Theorem imposes no constraints on the order of summation and activation functions, GAT also adheres perfectly to the universal approximation theorem, enabling it to approximate the reweighting matrix $\mtx{A}$ with arbitrary precision.

\end{proof}

\subsection{Proof of \method{} Approximation} \label{appdix:theorem2}

\begin{restatable}[\method{} Approximation]{theorem}{MethodMap}
    Given a linear \method{} module with an infinite hidden dimension, \ie, 
    \begin{equation} \label{eq_appdix:method}
        \mtdfunc(\mtx{X}) = \mtx{X} + \operatorname{TDC}_{\uparrow} \left(\operatorname{GAT}\left(\operatorname{TDC}_{\downarrow}\left(\mtx{X}\right)\right)\right),
    \end{equation}
    then $\mtdfunc(\mtx{X})$ could approximate $\mtx{Y} = \mtx{A}\mtx{X}$ in Propositions~\ref{prop:gen_linear_map} with an arbitrary precision.
\end{restatable}

\begin{proof}
    When using a depthwise convolution with a stride of 1, zero-padding, and a convolution kernel where only the first element is 1 while all others are 0, the convolution operation automatically degenerates into the identity mapping $f(x)= x$. Under this circumstance, Eq. \eqref{eq_appdix:method} would be further simplified as
    \begin{align}
        \mtdfunc(\mtx{X}) = \mtx{X} + \operatorname{GAT} (\mtx{X})
    \end{align}
    Then, since Theorem \ref{thm:gat} has no requirement on the approximated matrix $\mtx{X}$, we could leverage a GAT layer with a infinite hidden dimension to approximate the matrix $\mtx{A} - \mtx{I}$ with an arbitrary precision. Therefore, the formula of \method{} would be expressed as
    \begin{equation}
        \begin{aligned}
            \mtdfunc(\mtx{X}) &= \mtx{X} + \operatorname{GAT} (\mtx{X}) \\
            &\approx \mtx{X} + (\mtx{A} - \mtx{I}) \mtx{X} = \mtx{A} \mtx{X}
        \end{aligned}
    \end{equation}
\end{proof}

\section{Training Objectives} \label{appdix:train_obj}

In this section, we aim to propose an effective unsupervised training strategy for \method{}.
Our training strategy is designed to meet three critical objectives: 
(1) enable the model to effectively extract features from target domain samples; (2) maintain strong classification capabilities; and (3) align feature distributions between the source and target domains. To meet these objectives, we introduce three distinct loss functions that collectively guide the effective training of \method{}. The third objective has already been proposed in Section \ref{sec:train}. Therefore, we only address the first two objectives here.

First, to improve the model's understanding of the target domain, prior UDA works \cite{he2023domain, ghifary2016deep, zhuo2017deep} often rely on reconstruction loss, which serves as an additional supervision for the classification on unlabeled target domain. Reconstruction loss ensures that the decoded output closely resembles the input on the target domain, requiring the encoded features to preserve all information from the target domain. However, the usage of reconstruction loss could be harmful to MTS classification. In MTS classification tasks, temporal properties, such as periodicity and trends, are more strongly correlated with the labels whereas local noise may be detrimental to classification performance. But reconstruction loss, by design, does not differentiate between meaningful features and noise, making it a suboptimal choice for such tasks.
To address this issue, we propose the use of forecasting loss as an alternative. Forecasting tasks inherently require the model to focus more on temporal properties like trends and periodicity, while ignoring local random fluctuations. Consequently, features extracted for forecasting naturally transfer well to the classification task.

However, directly applying the forecasting loss presents challenges since the forecasting task requires the time series data to be sliced into adjacent historical and future segments.
To address this, given a sample $\mtx{X}_i^t \in \mathbb{R}^{D \times T}$ from the target domain, we first slice the samples into overlapping time windows $\mathcal{W}^t_{i, k} = \mtx{X}_i^t[:, k:k+L], k \in \{1, \dots, T-L\}$ \footnote{We use $\mathbf{X}[:, t_1:t_2]$ to represent a sliced segment from time $t_1$ to time $t_2$ of $\mathbf{X}$. All the slicing notations follow Python standards.} with $L$ being the window length, and then use those sliced time windows as the training inputs. Specifically, given a model integrated with \method{}, we leverage all the blocks without the last classification head as the feature extractor $\adpf(\cdot)$ and introduce a new forecasting projection head $\fctHead(\cdot)$. Then the forecasting loss could be represented as:
\begin{equation}
    \begin{aligned}
        \mathcal{L}_{\mathtt{f}} = \sum_{i=1}^{n_t} \sum_{k=1}^{T-2L} \operatorname{MAE}(\fctHead(\adpf(\mathcal{W}^t_{i, k})), \mathcal{W}^t_{i, k+L})
    \end{aligned}
\end{equation}
where $\operatorname{MAE}$ represent the mean absolute error. Here, $\mathcal{W}_{i, k}$ and $\mathcal{W}_{i, k+L}$ are two adjacent time windows, indicating the history information and future, respectively.

Second, to ensure that \method{} maintains the classification capabilities of the pretrained model, we perform a classification task using the labeled data from the source domain. To make the temporal dimension align with the previous forecasting task, we calculate the cross-entropy loss for each sliced time window:
\begin{equation} \label{eq:cls_loss}
    \mathcal{L}_{\mathtt{c}} = \sum_{i=1}^{n_s} \sum_{k=1}^{T-L} \ell_{\texttt{CE}} (g_\mathtt{c}(\adpf(\mathcal{W}^s_{i,k})), y_i^s)
\end{equation}
where $\ell_{\texttt{CE}}$ is  the cross-entropy loss function, $g_\mathtt{c}$ is the classification head, and $\mathcal{W}^s_{i,k}$ is the sliced time window from the source domain. During the inference phase, we randomly sample $m$ time windows from the entire time series and use a majority voting scheme to predict the label for the entire sequence based on the predictions from the sliced windows.

To sum up, we combine the classification loss $\mathcal{L}_\mathtt{c}$ on the source domain, the forecasting loss $\mathcal{L}_\mathtt{f}$ on the target domain, and the correlation alignment loss $\mathcal{L}_\mathtt{corr}$ across these two domains to formulate the final loss function. This unified objective ensures that the model not only learns discriminative features for classification but also captures temporal properties and reduces correlation shift effectively. Mathematically, the loss function can be expressed as:
\begin{equation}
    \mathcal{L} = \mathcal{L}_\mathtt{c} + \lambda_\mathtt{f} \mathcal{L}_\mathtt{f} + \lambda_\mathtt{corr} \mathcal{L}_\mathtt{corr}
\end{equation}
where $\lambda_\mathtt{fct}$ and $\lambda_\mathtt{cor}$ are two hyperparameters

\section{Dataset Description} \label{appdix:dataset}
\begin{table}[htbp]
    \centering
    \caption{The statistics of datasets.}
    \resizebox{0.5\linewidth}{!}{
    \begin{tabular}{c|cccc}
        \toprule
         Dataset & \# Domains & \# Timestamps & \# Variables \\
        \midrule
         HAR & 30 & 128 & 9 \\
         WIDSM & 36 & 128 & 3 \\
         HHAR & 9 & 128 & 3 \\
         Boiler & 3 & 128 & 20\\
        \bottomrule
    \end{tabular}
    }
    \label{tab:data_stat}
\end{table}
In this paper, we validate the effectiveness of \method{} on four different datasets, HAR \cite{anguita2013public}, WISDM \cite{wisdm_smartphone_and_smartwatch_activity_and_biometrics_dataset__507}, HHAR \cite{stisen2015smart}, and Boiler \cite{cai2021time}. The statistics of datasets are provided in Table \ref{tab:data_stat}, and the detailed information is listed below.

\begin{itemize}
    \item \textbf{HAR dataset}. The Human Activity Recognition Dataset has been collected from 30 subjects performing six different activities (Walking, Walking Upstairs, Walking Downstairs, Sitting, Standing, Laying). It consists of inertial sensor data that was collected using a smartphone carried by the subjects.
    \item \textbf{WISDM dataset}. WISDM Smartphone and Smartwatch Activity and Biometrics Dataset collects raw accelerometer and gyroscope sensor data from the smartphone and smartwatch at a rate of 20Hz. It is collected from 51 test subjects as they perform 18 activities for 3 minutes apiece. 
    \item \textbf{HHAR dataset}. The Heterogeneity Dataset for Human Activity Recognition contains the readings of two motion sensors commonly found in smartphones. Reading were recorded while nine users executed six different activities scripted in no specific order carrying smartwatches and smartphones.
    \item \textbf{Boiler dataset}. The boiler data consists of sensor data from three boilers from 2014/3/24 to 2016/11/30. There are 3 boilers in this dataset and each boiler is considered as one domain. We slice the original time series data with a time window of 128 and a stride of 32.
\end{itemize}

\section{Domain Pair Selection} \label{appdix:domain_select}
In this study, we utilize four datasets, each containing a large number of domains. As a result, exhaustively evaluating all possible source-target domain pairs is impractical (for example, 900 pairs for the HAR dataset). Therefore, selecting reasonable and effective source-target domain pairs becomes critically important.

To address this, we adopt the following domain pair selection mechanism: For each source-target domain pair, we compute the Wasserstein distance between samples sharing the same label in the source and target domains. We then sum the distances across all possible labels. Mathematically, this distance can be expressed as:
\begin{equation}
    \begin{aligned}
        d = \sum_{y \in \mathcal{Y}} \operatorname{Wass}(\mathcal{P}_S^y, \mathcal{P}_T^y)
    \end{aligned}
\end{equation}
where $\mathcal{P}_S^y$ and $\mathcal{P}_T^y$ represent the distributions of samples with label $y$ in the source domain $S$ and target domain $T$, respectively, and $\operatorname{Wass}(\cdot,\cdot)$ denotes the Wasserstein distance. This distance $d$ quantifies the similarity between the source and target domains: the smaller the distance, the smaller the domain shift, and the lower the difficulty of domain adaptation.

For HAR, HHAR and WISDM datasets, we divide all domain pairs into 10 groups, sorted by increasing the distance $d$. From each group, we sample one domain pair. This strategy ensures that the selected domain pairs represent varying levels of domain adaptation difficulty, from small to large domain shifts. For the Boiler dataset, due to its quite limited domain pairs (3 domains and 6 domain pairs in total), we only choose the domain pair with the largest $d$ and the smallest $d$, respectively.

The experimental results, summarized in Table \ref{tab:comprehensive_compare}, demonstrate the performance of our method across these selected domain pairs. Note that within each dataset, the domain pairs from the top to the bottom in Table \ref{tab:comprehensive_compare} are ordered by increasing $d$, indicating progressively higher domain adaptation difficulty (e.g., in the HAR dataset, the pair \mapping{24}{27} represents the smallest difficulty, while \mapping{19}{10} represents the largest difficulty).

\section{Description of Baselines}\label{appdix:baseline}
In this paper, we compare \method{} with 5 different baselines. These baselines could be roughly divided into three different categories. 

First, correlation-related UDA method is
\begin{itemize}
    \item \textbf{CORAL} \cite{sun2016deep} learn a nonlinear transformation that aligns correlations of layer activations in deep neural networks.
\end{itemize}

Second, MTS-related UDA methods include
\begin{itemize}
    \item \textbf{Raincoat} \cite{he2023domain} uses time and frequency-based encoders on the polar coordinate of frequency to learn domain-invariant time series representations.
    \item \textbf{SASA} \cite{cai2021time} introduces the intra-variables and inter-variables sparse attention mechanisms to extract associative structure time-series data with considering time lags for domain adaptation.
    \item \textbf{CLUDA} \cite{ozyurt2022contrastive} proposes a contrastive learning framework to learn domain-invariant, contextual representation for UDA of time series data.
\end{itemize}

Third, we introduce an adapter-related UDA method:
\begin{itemize}
    \item \textbf{UDApter} \cite{malik2023udapter} adds a domain adapter to learn domain-invariant information and a task adapter that uses domain-invariant information to learn task representations in the source domain.
\end{itemize}

\section{Step-by-step Incremental Adjustment} \label{appdix:incremental_adjust}
In the ablation study, we progressively adjusted the vanilla Transformer to the \method-enhanced Transformer, resulting in a significant improvement in accuracy from 79.79\% to 98.98\%. Specifically, we introduced the following six incremental adjustments:

\begin{enumerate}
    \item \textbf{+ Adapter (Eq. 1).} We incorporate the adapter defined in Eq. \eqref{eq:typ_adp} into the vanilla Transformer and trained it using the classification loss function $\mathcal{L}_\mathtt{c}$ in Eq. \ref{eq:cls_loss} on the source domain. This modification results in an accuracy improvement of 1.01\%.
    \item \textbf{+ correlation loss.} We optimize the adapter using a combination of classification loss and correlation alignment loss. This step further enhances accuracy by 9.89\%.
    \item \textbf{Adapter $\rightarrow$ \method{}.} We replace the adapter in Eq. \ref{eq:typ_adp} with \method{} and train it with the combined classification and correlation alignment loss. This substitution improved accuracy by 2.02\%.
    \item \textbf{+ window slicing.} To align with the setting of forecasting loss, we slice the original samples with a length of 128 into overlapping time windows with a length of 48 and used these sliced windows as inputs to train \method{}. This adjustment yields an additional accuracy gain of 2.89\%.
    \item \textbf{forecasting loss.} We introduce the forecasting loss, which uses consecutive time windows as input and their corresponding ground truth for prediction. The final loss function $\mathcal{L}$ in Eq. \eqref{eq:final_loss_func} is then leveraged to train \method{}, resulting in an accuracy improvement of 0.8\%.
    \item \textbf{+ max voting.} We apply a max-voting method to assign the label of the original sample based on predictions from its sliced time windows. This final step further boosted accuracy by 3.32\%.
\end{enumerate}

\section{Implementation Details} \label{appdix:implemetation}
\begin{table}[h]
    \centering
    \caption{Hyperparameters of backbone models.}
    \label{tab:hyperparam_backbone}
    \resizebox{0.7\linewidth}{!}{
    \begin{tabular}{c|ccccc}
    \toprule
        Hyperparameter & e\_layers & d\_model & d\_ff & top\_k & epoch (pretrain) \\
        Value & 3 & 128 & 256 & 3 & 10 \\
    \bottomrule
    \end{tabular}
    }
\end{table}
We use the code from Time-Series-Library repository \footnote{\url{https://github.com/thuml/Time-Series-Library}} to construct three different Transformer variants as backbone models, Transformer, TimesNet, and iTransformer. The hyperparameters for these three models follow the default configuration on Time-Series-Library repository, as shown in Table \ref{tab:hyperparam_backbone}. For \method, we use the TCNs with a kernel size $r=5$ and a padding of 2. We use Xavier initialization for the down-project TDC and GAT, and zero initialization for the up-down TDC. For training, we set the length of sliced time windows as 48 and set the number of sampled windows $m$ for max voting as 16. We use Adam optimizer with a learning rate of 1e-4, and set $\lambda_{\mathtt{c}}=0.5$ and $\lambda_{\mathtt{f}}=0.5$. 

\section{Comparison Between Correlation Alignment Loss and CORAL Loss} \label{appdix:compare_coral}

CORAL loss \cite{sun2016deep} is one widely-used domain adaptation loss, which focuses on minimizing the covariance between the source samples and the target samples.
In this section, we will demonstrate that the correlation alignment loss offers advantages over the CORAL loss. Furthermore, we show that under certain simplified conditions, the correlation alignment loss can be reduced to the CORAL loss, providing a unified perspective on both approaches. 

Our correlation alignment loss aim to use MMD to minimize the mean of the distributions of $\cor\left(\mtx{H}^s\right)$ and $\cor\left(\mtx{H}^t\right)$. Let the distributions  of $\cor\left(\mtx{H}^s\right)$ and $\cor\left(\mtx{H}^t\right)$ be denoted as $\mathcal{C}^s$ and $\mathcal{C}^t$, respectively. Mathematically, we aim to optimize the following equation.
\begin{equation}
    \begin{aligned}
        \mathcal{L}_\mathtt{corr} &=\operatorname{MMD} (\mathcal{C}^s, \mathcal{C}^t) \\
        &= \left\Vert \mathbb{E} \left[\psi\left(\cor\left(\mtx{H}^s\right)\right)\right] - \mathbb{E} \left[\psi\left(\cor\left(\mtx{H}^t\right)\right) \right]\right\Vert_2
    \end{aligned}
\end{equation}
where $\psi(\cdot)$ is one feature mapping function and $\operatorname{corr}(\mtx{H}) = \operatorname{vec} \left(\frac{\left(\mtx{H}\right)\left(\mtx{H} \right)^T}{\Vert \mtx{H} \Vert^2_\mathcal{F}}\right)$. Here, let us relax this feature mapping function to be the identity function, \ie, $\psi(\mtx{X}) = \mtx{X}$. Then our optimization objective could be further deduced:
\begin{equation}
\begin{aligned}
    \mathcal{L}_\mathtt{corr} 
    & = \left\Vert \mathbb{E} \left[\cor\left(\mtx{H}^s\right)\right] - \mathbb{E} \left[\cor\left(\mtx{H}^t\right)\right]\right\Vert_2 \\
    & = \left\Vert \mathbb{E}\left[\hat{\mtx{h}}^s (\hat{\mtx{h}}^s)^T \right] - \mathbb{E}\left[\hat{\mtx{h}}^t (\hat{\mtx{h}}^t)^T \right] \right\Vert_2 \\
    & = \left\Vert \mathbb{E} \left[ \left( \hat{\mtx{h}}^s - \mathbb{E}\left[\hat{\mtx{h}}^s\right]\right) \left( \hat{\mtx{h}}^s - \mathbb{E}\left[\hat{\mtx{h}}^s\right]\right)^T\right] \right. \\
    & \quad - \mathbb{E} \left[ \left( \hat{\mtx{h}}^t - \mathbb{E}\left[\hat{\mtx{h}}^t\right]\right) \left( \hat{\mtx{h}}^t - \mathbb{E}\left[\hat{\mtx{h}}^t\right]\right)^T\right] \\
    &\left. \quad + \mathbb{E} \left[\hat{\mtx{h}}^s \left(\hat{\mtx{h}}^s\right)^T \right] - \mathbb{E} \left[\hat{\mtx{h}}^t \left(\hat{\mtx{h}}^t\right)^T \right] \right\Vert_2 \\
\end{aligned}
\end{equation}
where $\hat{\mtx{h}}^s$ and $\hat{\mtx{h}}^t$ are the normalized vector from $\operatorname{vec}(\mtx{H}^s)$ and $\operatorname{vec}(\mtx{H}^t)$, \ie, $\hat{\mtx{h}}^s = \frac{\operatorname{vec}\left(\mtx{H}^s\right)}{\Vert \operatorname{vec}\left(\mtx{H}^s\right) \Vert_2}$ and  $\hat{\mtx{h}}^t = \frac{\operatorname{vec}\left(\mtx{H}^t\right)}{\Vert\operatorname{vec}\left(\mtx{H}^t\right) \Vert_2}$. 
Due to the triangle inequality, we have 
\begin{equation}
\begin{aligned}
    & \mathcal{L}_{\mathtt{corr}} \leq \mathcal{L}_{\mathtt{CORAL}} + \mathcal{L}_\mathtt{mean}, \\
    \text{where} \  &  \mathcal{L}_{\mathtt{CORAL}} = \left\Vert \mathbb{E} \left[ \left( \hat{\mtx{h}}^s - \mathbb{E}\left[\hat{\mtx{h}}^s\right]\right) \left( \hat{\mtx{h}}^s - \mathbb{E}\left[\hat{\mtx{h}}^s\right]\right)^T\right] 
     - \mathbb{E} \left[ \left( \hat{\mtx{h}}^t - \mathbb{E}\left[\hat{\mtx{h}}^t\right]\right) \left( \hat{\mtx{h}}^t - \mathbb{E}\left[\hat{\mtx{h}}^t\right]\right)^T\right] \right\Vert_2 \\
    \text{and} \ & \mathcal{L}_\mathtt{mean} = \left\Vert \mathbb{E} \left[\hat{\mtx{h}}^s \left(\hat{\mtx{h}}^s\right)^T \right] - \mathbb{E} \left[\hat{\mtx{h}}^t \left(\hat{\mtx{h}}^t\right)^T \right] \right\Vert_2
\end{aligned}
\end{equation}
Here, $\mathcal{L}_\mathtt{CORAL}$ represents the original loss proposed by CORAL \cite{sun2016deep}, and $\mathcal{L}_\mathtt{mean}$ minimizes the discrepancy between the mean distributions of the source and target domains. Thus, the correlation alignment loss not only aligns the multivariate correlation between the source and target domains, as CORAL does, but also reduces the mean differences between the two domains.

Compared to CORAL and its following works, the correlation alignment loss simultaneously supervises both covariance and mean alignment, ensuring more precise domain alignment. Notably, when the mean distributions of the source and target domains coincide, the correlation alignment loss naturally reduces to the CORAL loss.
\end{document}